\documentclass[11pt, letterpaper]{article}
\usepackage[margin=1.0in]{geometry} 
\usepackage[utf8]{inputenc}

\usepackage{nicefrac}
\usepackage{amssymb}
\usepackage{amsmath}
\usepackage{hyperref}
\usepackage{cleveref}
\usepackage{xurl}
\newcommand{\LeftEqNo}{\let\veqno\@@leqno}
\usepackage{environ}
\usepackage{xcolor}
\usepackage{float}
\usepackage{adjustbox}
\usepackage{graphicx}
\usepackage{multirow}
\usepackage{makecell}
\usepackage{amsmath}
\usepackage{amssymb}
\usepackage[scr=kp]{mathalpha}
\usepackage{braket}
\usepackage[noend]{algpseudocode}
\usepackage{algorithm}
\usepackage{tikz}
\usetikzlibrary{positioning,calc} 
\usepackage{pgfplots}
\usepackage{soul}
\usetikzlibrary{arrows,decorations.markings}
\usepackage{float}
\usepackage{caption}
\usepackage{graphicx}
\usepackage{array}
\usepackage{enumerate}
\usepackage{booktabs,caption,threeparttable, multirow}
\usepackage{siunitx}
\let\tnote\relax
\usepackage{subfigure}
\usepackage{comment}
\usepackage{lscape}
\pgfplotsset{compat=1.17} 
\usepackage{titlesec}
\usepackage[title]{appendix}

\usepackage{natbib}
\providecommand{\keywords}[1]
{
  \small	
  \textbf{Keywords:} #1
}

\title{Constraint Learning to Define Trust Regions in Predictive-Model Embedded Optimization} 

\author{Chenbo Shi\footnote{Department of Operations and Information Management, School of Business, University of Connecticut, chenbo.shi@uconn.edu}, Mohsen Emadikhiav\footnote{Department of Information Technology and Operations Management, Florida Atlantic University, memadikhiav@fau.edu}, Leonardo Lozano\footnote{epartment of Operations, Business Analytics, and Information Systems, University of Cincinnati, leolozano@uc.edu}, David Bergman\footnote{Department of Operations and Information Management, School of Business, University of Connecticut, david.bergman@uconn.edu}}

\date{\today}

\begin{document}

\maketitle

\begin{abstract}
There is a recent proliferation of research on the integration of machine learning and optimization. One expansive area within this research stream is predictive-model embedded optimization, which proposes the use of pre-trained predictive models as surrogates for uncertain or highly complex objective functions. In this setting, features of the predictive models become decision variables in the optimization problem. Despite a recent surge in publications in this area, only a few papers note the importance of incorporating trust region considerations in this decision-making pipeline, i.e., enforcing solutions to be similar to the data used to train the predictive models. Without such constraints, the evaluation of the predictive model at solutions obtained from optimization cannot be trusted and the practicality of the solutions may be unreasonable. In this paper, we provide an overview of the  approaches appearing in the literature to construct a trust region, and propose three alternative approaches. Our numerical evaluation highlights that trust-region constraints learned through isolation forests, one of the newly proposed approaches, outperform all previously suggested approaches, both in terms of solution quality and computational time.    
\end{abstract}


\section{Introduction} \label{sec:introduction}
The use of pre-trained predictive models in optimization decision problems is nothing new. In this setting, the collected data includes a set of controllable independent variables as past decisions (sometimes together with uncontrollable variables) and their corresponding outcomes. Practitioners benefit by learning from data to make decisions that result in more favorable outcomes \citep{biggs2017optimizing, mivsic2020optimization} through surrogate models, to learn the underlying relationship between the controllable independent variables and their outcome \citep{biggs2017optimizing, grimstad2019relu}. Once the relationship is learned, one can solve an optimization problem defined over the predictive model, where decision variables in the optimization problem correspond to the controllable independent variables, and the objective function is given by the outcome of the pre-trained predictive model. We will refer to this decision-making paradigm as \emph{predictive-model embedded optimization} (PMO).
 
Examples abound; consider a traditional setting in advertising. Suppose a retailer can select from among $n$ advertising channels.  Given a budget of $B$ dollars, how much should the retailer allocate to each advertising channel to maximize revenue? Classical Operations Research methods suggest that a predictive model such as regression model can be fit on historical data (including both previous decisions, market conditions, and revenue) to predict revenue, and then the allocation decision variables can be selected to maximize predicted revenue. Depending on the predictive model chosen, the optimization model can be appropriately formulated. For example, if the predictive model chosen is a linear regression, the optimization problem can be modeled as a linear programming model; if the predictive model chosen is a neural network or a random forest, recent research shows how a mixed-integer model can be formulated (for instance, \cite{anderson2020strong}).

Because of its generality, PMO finds potential applications throughout the analytical landscape. There is a major issue with the real-world application of PMO, which practitioners are well aware of: decisions produced by PMO are often unrealistic because they can be very different from what is observed historically in data. This happens because often the existing PMO frameworks ignore the \textit{trust region}, i.e., ensuring that solutions to the optimization problem are similar to the data used to train the predictive models \citep{maragno2021mixed, mistry2021mixed, schweidtmann2022obey}.

Let us demonstrate this issue using an example in a wine production context on a dataset that has been used before in the PMO literature \citep{mivsic2020optimization, wang2021two}. Wine scores are typically used by critics to specify their opinion about wine quality. Suppose a wine producer is equipped with a dataset that includes physicochemical property measurements of varieties of wine as features (e.g., pH, fixed acidity, residual sugar, etc.) along with the corresponding wine scores. The producer wants to leverage this dataset and build a PMO model to determine the best combination of physicochemical property measurements that maximizes the expected wine score. 

\begin{figure}[h!]
	\centering
	\captionsetup{width=\linewidth}
	\includegraphics[scale=0.4]{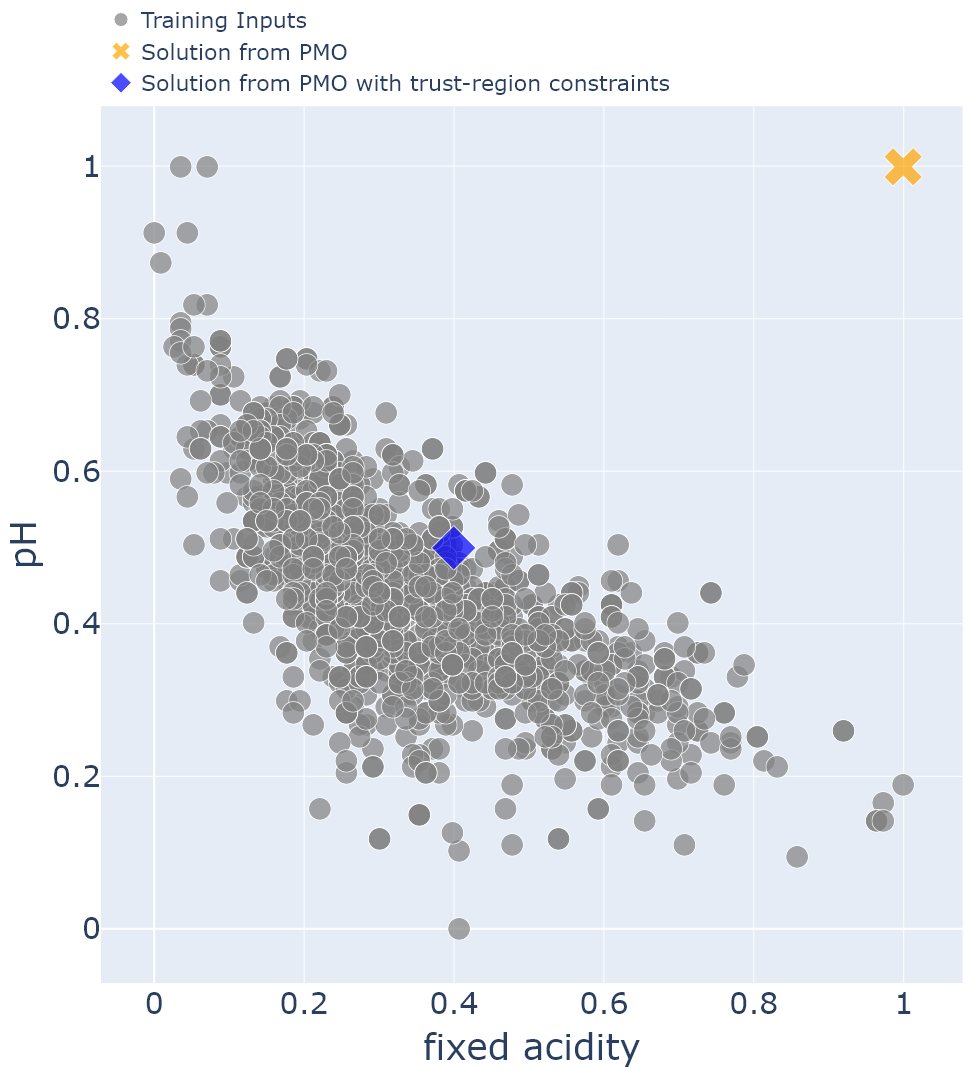}
	\caption{A scatter plot comprising two variables (pH and fixed acidity) for two solutions from a PMO model}
	\label{fig:unrestricted solution from wine data}
 \end{figure}

Figure \ref{fig:unrestricted solution from wine data} shows a scatter plot (pH versus fixed acidity) for two solutions obtained via a PMO model defined over a trained neural network (which is one of the most popular predictive models nowadays) and historical observations in the data (features are scaled between 0 and 1). The first solution is obtained by solving the corresponding optimization model without additional constraints (represented by an orange cross) and the second solution is obtained by incorporating trust-region constraints (represented by a blue diamond). Measurements of pH and fixed acidity in historical observations present a negative correlation, which is expected as higher acidity should correspond with a lower pH. The initial solution obtained, however, completely violates this relationship, rendering the solution unacceptable. Alternatively, the second solution is closer to the training data and aligns with the aforementioned chemical considerations. It may be argued that the predictive model is not good enough to capture the inverse correlation. However, in predictive modeling applications, a predictive model is usually selected based on its overall predictive performance instead of considering such specific correlation relationships between pairs of features. By solving an optimization problem defined over such a predictive model, we cannot guarantee that the solution obey the practical requirements. In fact, it is a common character of solutions of PMOs falling at the extremes of the feasible region \citep{maragno2021mixed}. 
 
The example above highlights how the solutions obtained by PMO can be inappropriate for a particular application. That is not the only potential issue. Even if a wine could be created that has the relationship of high acidity and high pH, the evaluation of the objective function of the pre-trained model at that point would be untrustworthy. If the optimization problem selects a solution that is too different from the training data of the predictive model, one cannot trust the prediction produced by the predictive model at that point (i.e., the error is immeasurable). This is well known in the application of predictive modeling, where it is common to assume that the unobserved/test data comes from the same distribution as the historical/training data.

There are therefore two major issues with a direct application of PMO as presented thus far in this paper: (1) the solutions can be inappropriate for the application, and (2) the validity of the predictive model at the solution obtained could be untrustworthy. To address these issues, constraints can be added for each possible  application to ensure that the decisions model physical or other real-world considerations, but these relationships might be highly non-linear, or perhaps unknown to the modeler. One might also try to examine all pairs of variables and make sure they are reasonably correlated as desired, but this can be challenging to identify for higher degree relationships and can lead to over constraining. 

Recently, some research has focused on developing methods to incorporate trust region in PMOs. For instance, \cite{biggs2017optimizing} and \cite{maragno2021mixed} use the convex-hull of observations in the dataset to model a trust region, while \citet{schweidtmann2022obey} utilize a one-class support vector machine (SVM) to determine constraints for the trust region. In essence, constructing a trust region for PMOs ensures that the solutions are not outliers compared to the historical observations. In this paper, we provide an overview of the existing methods for modeling the trust region in PMOs and explore three additional ways to detect outliers, which are melded into PMOs as a trust region: the first defines a trust region by selecting queries from a pre-trained \emph{isolation forest} \citep{liu2008isolation}; the second uses \emph{Mahalanobis distance} \citep{mahalanobis1936generalized} to constrain the solutions within a given distance of the training data, which is particularly useful when an assumption on multivariate normality is reasonable; and the third incorporates constraints to ensure that the solution is within a limited distance of \textit{K}-nearest neighbors \citep{altman1992introduction} in the training data.

We design computational experiments to compare the performance of our proposed trust-region constraints to the existing approaches in the literature. To our knowledge we are the first to evaluate different trust regions against each other both in terms of solution quality and computational performance (measured in solution resolution time). For our experiments, we generate a testbed of problem instances based on 7 different benchmark functions and use multiple predictive models to fit the training data within our PMO approach. Our results indicate that our proposed constraints compare favorably with the existing approaches from the literature, particularly our Isolation-Forest constraints. 

Our contributions are therefore as follows: 
\begin{enumerate}
    \item Although researchers have independently developed methods for modeling the trust region in PMOs, to our best knowledge, we are the first to provide a comparison of existing methods in terms of solution quality and CPU time.
    \item We propose three alternative methods for modeling the trust region in PMOs. Our results show the effectiveness of our proposed approaches, particularly the trust region defined by Isolation Forest, compared to existing ones in the literature both in terms of solution quality and computational times.
    \item We present a set of benchmark instances that can be used by the optimization community for further research on related topics.
\end{enumerate}

\section{Literature Review} \label{sec:literature}

Multiple works from the literature consider optimization problems defined over trained predictive models, ranging from simple linear regressions to more complex predictive models such as random forest and neural networks. \cite{bertsimas2016analytics} optimize over an estimated ridge regression model seeking to improve the efficacy of chemotherapy regimens in clinical trials. \cite{huang2019predictive} and \cite{baardman2019scheduling} propose optimization models defined over linear regressions in the context of retailers location decisions and scheduling of promotion vehicles, respectively. \cite{ferreira2016analytics} use random forest to estimate the demand of a product based on its price, and develop an optimization model over a trained random forest to determine product prices that maximize total expected revenue. Other application areas include food delivery \cite{liu2020}, scholarship allocation \cite{bergman2022janos}, personalized pricing \cite{biggs2021}, and auctions \cite{verwer2017}.   

Optimization problems defined over trained predictive models are often challenging to solve, depending on the properties of the underlying predictive model. \cite{mivsic2020optimization} and \cite{biggs2017optimizing} propose exact solution approaches for optimization problems defined over tree ensembles. Mixed integer programming (MIP) formulations for optimization models defined over trained neural networks are studied by \cite{cheng_maximum_2017, fischetti2018deep, bunel_unified_2018, dutta_output_2018, bergman2022janos, grimstad2019relu, schweidtmann2019deterministic, tjeng_evaluating_2019, grimstad_relu_2019, botoeva_efficient_2020, anderson2020strong} and \citet{tsay2021}. Ensembles of neural networks are studied by \cite{wang2021two}. A closely related research area deals with so-called \emph{verification problems}, which seek to verify relationships between the inputs and outputs of neural networks, as well as to measure the sensitivity of the outputs to small changes in the inputs. These problems are often formulated as mixed-integer programs defined over trained neural networks \citep{lomuscio_approach_2017,cheng_maximum_2017,katz2017reluplex,serra2018bounding}.

In predictive modeling, it is common to assume that the unobserved/test data comes from the same distribution as the historical/training data, which is used to build predictive models \citep{amodei2016concrete}, otherwise the predictive models' applicability would be limited. Similarly, a solution obtained via PMO can be seen as a new observation in the dataset. The prediction provided by the predictive model should only be trusted if the solution comes from the same distribution as the training data. To constrain the predictive models from extrapolating, a few recent studies have considered the notion of trust region in PMOs. \cite{teixeira2006bioprocess} use a clustering technique to define a trust region of a nonparametric input space where the predictions of the predictive model are reliable. \cite{biggs2017optimizing} and \cite{ maragno2021mixed} define a similar subspace using the convex hull of observations in the training data to model a trust region by constraining PMO solutions to be within this convex hull. \cite{mistry2021mixed} add a penalty term to the objective function so that solutions that lie far away from the subspace constructed by principle component analysis (PCA), are penalized. \cite{schweidtmann2022obey} constrain the solutions to belong to a validity domain defined by a pre-trained one-class SVM, a classifier for outlier detection.


\section{PMO with Trust-Region Constraints} \label{sec:models}

We consider a trained predictive model that is established to learn the relationship between a dependent variable $y$ and a vector of independent variables $\boldsymbol{x}=\{x_1,..., x_n\}$ based upon a set of historical observations $\mathcal{S}=\{(\boldsymbol{\hat{x}}^j, \hat{y}^j)\}_{j=1}^q$. We refer to set $\mathcal{S}^x=\{\boldsymbol{\hat{x}}^j \}_{j=1}^q$ as the training input data of the problem and refer to the components of vector $\boldsymbol{x}$ as features with index set given by $\mathcal{I} = \set{1,\dots,n}$. For simplicity, we assume that all the independent variables are continuous and each independent variable is scaled to $[0, 1]$. We define decision variables $\boldsymbol{x}$ to determine the value of each independent variable and let the predicted outcome of the predictive model be given by $F(\boldsymbol{x})$. We study problems of the following form:
\begin{subequations}
\begin{align}
\min_{\boldsymbol{x}}  \quad & F(\boldsymbol{x})  \label{model:PMO}\\
\textrm{s.t.} \quad & \boldsymbol{x} \in \mathbf{[0,1]}.  \label{constr:bound}
\end{align}
\end{subequations}

We consider three different predictive models associated with function $F(\boldsymbol{x})$: linear regression (LR), random forest (RF), and neural networks (NN). The mathematical formulations corresponding to each one of the three predictive models are discussed in Appendix \ref{sec:appendix1}. 
 Constraints are then added to the model above to ensure that solutions found are within a trust region. In Section \ref{sec:overview of existing tr constraints} we first review three existing types of trust region constraints from the literature. We then introduce our proposed constraints in Section \ref{sec:our tr constraints}.

\subsection{Overview of the Existing Trust-Region Constraints} \label{sec:overview of existing tr constraints}
\subsubsection{Convex Hull Constraints} \label{subsec:PMO-CH model} \hfill \break
 \cite{maragno2021mixed} characterize a trust region using the convex hull (CH) of the training input data $\mathcal{S}^x$, which is the smallest convex polytope that contains all the points in $\mathcal{S}^x$. For each sample point $\boldsymbol{\hat{x}}^j$ in $\mathcal{S}^x$, an auxiliary decision variable $w_j$ is added to describe the convex hull. Let $\mathcal{J}=\{1,...,q\}$ be the index set of $\mathcal{S}^x$. The trust region is then represented by the following constraints:
\begin{subequations}
\begin{align}
& \boldsymbol{x} = \sum_{j \in \mathcal{J}} w_j \boldsymbol{\hat{x}}^j  \label{constraint:CH1}\\
& \sum_{j \in \mathcal{J}} w_j = 1 \label{constraint:CH2}\\
& w_j \geq 0  & \forall j \in \mathcal{J}, \label{constraint:CH3}
\end{align}      
\end{subequations}
which ensure that any PMO solution belongs to the convex hull of the training input data.

\subsubsection{One-Class SVM Constraints} \label{subsec:PMO-SVM model} \hfill \break
One-class SVM is an unsupervised machine learning algorithm that specifies a boundary to classify sample points for outlier detection. In other words, if a sample point falls outside of the boundary learned through one-class SVM, it is classified as an outlier. 
The boundary is mathematically represented using a \textit{decision function} which may take various forms. The form of the decision function depends on the type of \textit{kernel} function used to model the data. In SVMs, a kernel is a function that is used to transform the data into higher dimensions so that data points can be linearly separable \citep{hofmann2008kernel}. Depending on the complexity of the data, different kernel functions such as linear, polynomial, and Gaussian, among others may be used. It has been shown that complex boundaries can be accurately modeled using a Gaussian radial basis kernel \citep{tax2001one}. Let $I_{SV} \subseteq \mathcal{S}^x$ denote the set of support vectors from the training input data. A decision function based on a Gaussian radial basis kernel is:  
\begin{equation}
\sum_{j \in I_{SV}} \alpha_j \text{exp}(-\gamma ||\boldsymbol{\hat{x}}^j - \boldsymbol{x}||^2) - \rho,    
\end{equation}
where hyper-parameter $\gamma$ is tuned before the training process, while parameters $\alpha$ and $\rho$ are learned during the training process for one-class SVMs \citep{scholkopf1999support}. 

\cite{schweidtmann2022obey} propose trust-region constraints based on one-class SVM classifiers as the ones described above. These constraints ensure that PMO solutions $\boldsymbol{x}$ fall inside a trust region and are modeled as:
\begin{align}
& \sum_{j \in I_{SV}} \alpha_j \text{exp}(-\gamma ||\boldsymbol{\hat{x}}^j - \boldsymbol{x}||^2) \geq \rho, \label{constraint:SVM1}
\end{align}
which ensure that solutions $\boldsymbol{x}$ are not classified as outliers by the pre-trained one-class SVM classifier.

PMO models with trust region SVM constraints \eqref{constraint:SVM1} can be solved directly with off-the-shelf mixed-integer non-linear programming solvers such as BARON \citep{sahinidis:baron:21.1.13}. To boost the running time performance of this model, we alternatively propose a  heuristic branch-and-cut algorithm that enforces trust-region SVM constraints \eqref{constraint:SVM1} dynamically by checking if PMO solutions violate these constraints via callbacks. Our experiments show that even if the branch-and-cut algorithm is not able to prove optimality, it is able to find high quality solutions quickly. We present the details of this heuristic algorithm in Appendix \cref{appendix:appendix-svm-bc model}.

\subsubsection{Principal Component Analysis Constraints} \label{subsec:PMO-PCA model} \hfill \break
\cite{mistry2021mixed} incorporate the trust region into PMOs by penalizing solutions that are far away from the training input data. To fulfill this idea, they first define a subspace of the training input data that is created using principal component analysis \citep{jolliffe2002principal} and then include a convex penalty term to penalize the objective function based on the solution's distance to the defined subspace.

Specifically, given a training input data $\mathcal{S}^x$ with $n$ features, PCA defines a set of $n$ ordered and orthogonal principal components, $\boldsymbol{\Phi} = [\phi_1\ldots\phi_n]$. Among all the $n$ principal components, oftentimes the top $k$ (such that $k < n$) principal components, $\boldsymbol{\Phi'}=[\phi_1\ldots\phi_k]$, are able to explain most of the variance in $\mathcal{S}^x$. The resulting subspace of $\mathcal{S}^x$ is a vector denoted by $(\phi_1,\ \ldots,\ \phi_k)$. \cite{mistry2021mixed} use the following penalty term to penalize the solutions $\boldsymbol{x}$ that are far away from the desired subspace:
\begin{align}
\text{cvx}_{\lambda}\left(\boldsymbol{x}\right) = \lambda ||\left(\boldsymbol{I}-\boldsymbol{\Phi'}\boldsymbol{\Phi'}^T\right)\text{diag}\left(\boldsymbol{\sigma}\right)^{-1}\left(\boldsymbol{x}-\boldsymbol{\mu}\right)||_2^2, \label{constraint:PCA}
\end{align}
where $\lambda$ is a non-negative penalty parameter, $\boldsymbol{I}$ is the identity matrix, $\boldsymbol{\mu}$ is the sample mean and $\boldsymbol{\sigma}$ is the sample standard deviation. Note that PCA is calculated with standardized $\mathcal{S}^x$ since it is sensitive to scaling, and $\boldsymbol{x}$ is standardized using $\left(\boldsymbol{\sigma}\right)^{-1}\left(\boldsymbol{x}-\boldsymbol{\mu}\right)$ so that $\boldsymbol{x}$ and PCA are in the same scale.

In order to compare \cite{mistry2021mixed} penalty method of enforcing a trust region with the other constraint-based approaches surveyed, we model the trust region given by PCA with the following constraint:
\begin{align}
||\left(\boldsymbol{I}-\boldsymbol{\Phi'}\boldsymbol{\Phi'}^T\right)\text{diag}\left(\boldsymbol{\sigma}\right)^{-1}\left(\boldsymbol{x}-\boldsymbol{\mu}\right)||_2^2 \le \gamma, \label{constraint:PCA_constr}
\end{align}
where threshold $\gamma$ is a parameter that controls the looseness (tightness) of the constraint.

\subsection{Our Proposed Trust-Region Constraints} \label{sec:our tr constraints}
We propose three types of trust-region constraints based on three additional techniques used to identify outliers. 
\subsubsection{Isolation Forest Constraints} \label{subsec:PMO-IF model} \hfill \break
Our first proposed type of trust region constraints are originated from a pre-trained isolation forest classifier \citep{liu2008isolation}, which is a model used for outlier (anomaly) detection.

An isolation forest is an ensemble of \textit{isolation trees} where each isolation tree is trained over a randomly selected subsample of the training data. An isolation tree is a model that detects outliers using independent variables $\boldsymbol{x}$ by checking a series of \textit{splits}. A split specifies a condition on a single feature $i \in \mathcal{I}$ using a query of the form ``Is $x_i \le a$?", where $a$ is a random value within the domain of feature $i$. The series of splits arrange a tree, with each split node branching into two child nodes. To determine whether a data point is an outlier, we start from the root node of the tree and recursively check the query of each split node. If the condition is true, the data point is assigned to the left branch of the split node, otherwise it is assigned to the right branch of the split node. The data point eventually is assigned to a \textit{leaf node}, for which there are no more splits.

An isolation tree uses a predetermined threshold, $d$, associated with the depth of the leaf nodes. Data points that are assigned to leaf nodes with depth smaller than or equal to $d$ are classified as outliers. Figure \ref{fig:IF_sample_tree} illustrates an example of an isolation forest classifier having $d = 2$. Leaf nodes with depth smaller than or equal to 2 correspond to data points that are classified as outliers, while leaf nodes with depth larger than 2 correspond to data points that are classified as inliers.

\begin{table}[h!]
\begin{tabular}{c c c c}
    \begin{minipage}{.25 \linewidth}
        \begin{tikzpicture}
            \tikzstyle{level 1}=[sibling distance=24mm]
            \tikzstyle{level 2}=[sibling distance=12mm]
            \tikzstyle{level 3}=[sibling distance=6mm]
            \tikzstyle{level 4}=[sibling distance=3mm]
            \tikzset{
            splitnode/.style={circle,draw=red,fill=red,inner sep=1.5},
            outliernode/.style={rectangle,draw=blue,fill=blue,inner sep=2},
            inliernode/.style={rectangle,draw=blue,inner sep=2}
            }
            \node[splitnode, label=above:{tree 1}]{}
                child{node[splitnode]{}
                    child{node[outliernode]{}}
                    child{node[splitnode]{}
                        child{node[splitnode]{}
                            child{node[inliernode]{}}
                            child{node[inliernode]{}}
                        }
                        child{node[inliernode]{}}
                    }
                }
                child{node[splitnode]{}
                    child{node[splitnode]{}
                        child{node[splitnode]{}
                            child{node[inliernode]{}}
                            child{node[inliernode]{}}
                        }
                        child{node[splitnode]{}
                            child{node[inliernode]{}}
                            child{node[inliernode]{}}
                        }
                    }
                    child{node[outliernode]{}}
                }                
            ;
              \begin{scope}
                \draw [dashed] (-20mm,-34mm) -- (130mm,-34mm)node[above right]{$d=2$};
              \end{scope}
        \end{tikzpicture}        
    \end{minipage} %
    \begin{minipage}{.25 \linewidth}
        \begin{tikzpicture}
            \tikzstyle{level 1}=[sibling distance=24mm]
            \tikzstyle{level 2}=[sibling distance=12mm]
            \tikzstyle{level 3}=[sibling distance=6mm]
            \tikzstyle{level 4}=[sibling distance=3mm]
            \tikzset{
            splitnode/.style={circle,draw=red,fill=red,inner sep=1.5},
            outliernode/.style={rectangle,draw=blue,fill=blue,inner sep=2},
            inliernode/.style={rectangle,draw=blue,inner sep=2}
            }
            \node[splitnode, label=above:{tree $2$}]{}
                child{node[splitnode]{}
                    child{node[splitnode]{}
                        child{node[inliernode]{}}
                        child{node[splitnode]{}
                            child{node[inliernode]{}}
                            child{node[inliernode]{}}
                        }
                    }
                    child{node[splitnode]{}
                        child{node[inliernode]{}}
                        child{node[splitnode]{}
                            child{node[inliernode]{}}
                            child{node[inliernode]{}}
                        }
                    }
                }
                child{node[outliernode]{}}
            ;
        \end{tikzpicture}
    \end{minipage} %
    \begin{minipage}{.1 \linewidth}
        \ldots \ldots
    \end{minipage} %
    \begin{minipage}{.3 \linewidth}
        \begin{tikzpicture}
            \tikzstyle{level 1}=[sibling distance=24mm]
            \tikzstyle{level 2}=[sibling distance=12mm]
            \tikzstyle{level 3}=[sibling distance=6mm]
            \tikzstyle{level 4}=[sibling distance=3mm]
            \tikzset{
            splitnode/.style={circle,draw=red,fill=red,inner sep=1.5},
            outliernode/.style={rectangle,draw=blue,fill=blue,inner sep=2},
            inliernode/.style={rectangle,draw=blue,inner sep=2}
            }
            \node[splitnode, label=above:{tree \rule{.4pt}{1.6ex}$\, \mathcal{T}\, $\rule{.4pt}{1.6ex}}]{}
                child{node[splitnode]{}
                    child{node[splitnode]{}
                        child{node[splitnode]{}
                            child{node[inliernode]{}}
                            child{node[inliernode]{}}                        
                        }
                        child{node[splitnode]{}
                            child{node[inliernode]{}}
                            child{node[inliernode]{}}                        
                        }
                    }
                    child{node[splitnode]{}
                        child{node[splitnode]{}
                            child{node[inliernode]{}}
                            child{node[inliernode]{}}                        
                        }
                        child{node[splitnode]{}
                            child{node[inliernode]{}}
                            child{node[inliernode]{}}                        
                        }
                    }
                }
                child{node[splitnode]{}
                    child{node[splitnode]{}
                        child{node[splitnode]{}
                            child{node[inliernode]{}}
                            child{node[inliernode]{}}                        
                        }
                        child{node[splitnode]{}
                            child{node[inliernode]{}}
                            child{node[inliernode]{}}                        
                        }
                    }
                    child{node[splitnode]{}
                        child{node[splitnode]{}
                            child{node[inliernode]{}}
                            child{node[inliernode]{}}                        
                        }
                        child{node[splitnode]{}
                            child{node[inliernode]{}}
                            child{node[inliernode]{}}                        
                        }
                    }
                }
            ;
        \end{tikzpicture}
    \end{minipage}  
    \\
    \begin{minipage}{.9 \linewidth}
        \begin{tabular}{l l l}
            \tikz\draw[red,fill=red] (0,0) circle (.4ex); \text{split node} 
            & \tikz\draw [blue, fill=blue] (0.1,0.1) rectangle (0.25,0.25); \text{leaf node containing outliers} 
            & \tikz\draw [blue] (0.1,0.1) rectangle (0.25,0.25); \text{leaf node containing inliers} 
        \end{tabular}
    \end{minipage} 
\end{tabular} 
\captionof{figure}{Illustration of an isolation forest with threshold $d = 2$}
\label{fig:IF_sample_tree}
\end{table}




\begin{figure}[h!]
    \centering
    \captionsetup{width=\linewidth}
    \includegraphics[scale=0.5]{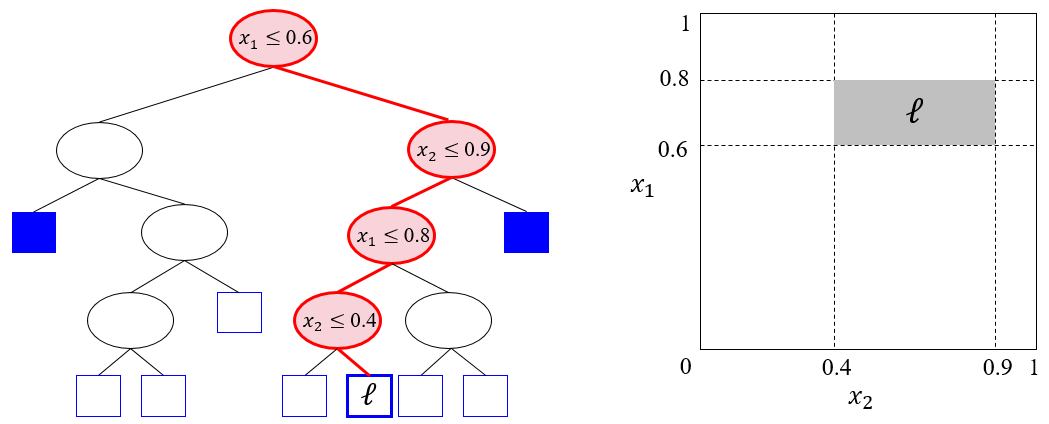}
    \caption{Left: tree $1$ from Figure \ref{fig:IF_sample_tree}, where the shaded split nodes and bold edges indicate how a solution that fulfills the queries in those split nodes falls into leaf node $\ell$. Right: corresponding domain of leaf node $\ell$, e.g., the lower bound in the dimension of feature 2 of leaf node $\ell$ from tree 1 is $l_{2,1,\ell} = 0.4$.}
    \label{fig:domain of a leaf node}
\end{figure}

We propose trust-region constraints to ensure that solutions $\boldsymbol{x}$ are not classified as outliers by a pre-trained isolation forest. The goal of these constraints obtained from an isolation forest is to ensure that PMO solutions do not correspond to any leaf node with a shallow depth, i.e., a depth less than or equal to $d$. We remark that any solution vector $\boldsymbol{x}$ corresponds to exactly one leaf node in each isolation tree and because there is a unique path between the root node of a tree and each leaf node, then leaf nodes correspond to a series of lower and upper bounds on the variable features, given by the queries associated with split nodes in the path from the root node. Figure \ref{fig:domain of a leaf node} illustrates the relationship between paths in an isolation forest and variable features bounds. We exploit these bounds to formulate our trust-region constraints as follows. 

Consider an isolation forest with $|\mathcal{T}|$ isolation trees. Let $\mathcal{L}_t$ be the set of leaf nodes in tree $t \in \mathcal{T}$. We introduce binary decision variable $y_{t, \ell}$ that takes a value of $1$ if leaf node $\ell \in \mathcal{L}_t$ from tree $t \in \mathcal{T}$ is selected. We introduce auxiliary decision variables $z_i^{LB}$ and $\ z_i^{UB}$ to specify the range of each variable feature $i \in \mathcal{I}$. We denote the depth of leaf nodes by $\delta(t, \ell)$, and let the lower and upper bounds corresponding to a leaf node be $l_{i,t,\ell}$ and $u_{i,t,\ell}$ for all $i \in \mathcal{I}$, $t \in \mathcal{T}$, and $\ell \in \mathcal{L}_t$. Our proposed isolation-forest constraints are:
\begin{subequations}
\begin{align}
& \sum_{\ell \in \mathcal{L}_t} y_{t, \ell} = 1 &\forall t \in \mathcal{T}  \label{constr:PMO-IF1}\\
&  y_{t, \ell} = 0  &\forall t \in \mathcal{T},\ \ell \in \mathcal{L}_t \ | \ \delta(t, \ell) \leq d  \label{constr:PMO-IF2}\\
& \sum_{\ell \in \mathcal{L}_t } l_{i,t,\ell} y_{t, \ell} \le z_i^{LB} &\forall t \in \mathcal{T},\  \forall i \in \mathcal{I} \label{constr:PMO-IF3}\\
& 1 - \sum_{\ell \in \mathcal{L}_t}(1 - u_{i,t,\ell})y_{t, \ell} \ge z_i^{UB} &\forall t \in \mathcal{T},\  \forall i \in \mathcal{I} \label{constr:PMO-IF4}\\
& z_i^{LB} \le x_i \le z_i^{UB}- \epsilon &\forall i \in \mathcal{I} \label{constr:PMO-IF6}\\
& z_i^{LB}, z_i^{UB} \in [0,1] &\forall i \in \mathcal{I} \label{constr:PMO-IF7}\\
& y_{t,\ell} \in \{0,1\} &\forall t \in \mathcal{T},\ \ell \in \mathcal{L}_t \label{constr:PMO-IF9}
\end{align}
\end{subequations}


Constraints \eqref{constr:PMO-IF1} ensure that exactly one of the leaf nodes from each tree in the isolation forest is selected. Constraints \eqref{constr:PMO-IF2} require that the depth of the selected leaf node $l$ in each tree $t$ is larger than the threshold $d$. Constraints \eqref{constr:PMO-IF3} and constraints \eqref{constr:PMO-IF4} ensure that, for any leaf node that is selected, the domain of the solution is limited to the feature ranges determined by the path leading to that leaf node. Constraints \eqref{constr:PMO-IF6} ensure that $\boldsymbol{x}$ belongs to the domain defined by the corresponding leaf nodes selected. The combination of these constraints ensure that solution $\boldsymbol{x}$ does not correspond to a leaf node of any tree that can be labeled as an outlier.

\subsubsection{Mahalanobis Distance Constraints} \label{subsec:PMO-MD model} \hfill \break
In Section \ref{subsec:PMO-PCA model}, we introduced a distance-based approach to penalizing solutions further away from the training data subspace defined by PCA. Instead of capturing the distance from a data point to a subspace, we propose to use the Mahalanobis Distance (MD)  \citep{mahalanobis1936generalized}, which measures the number of standard deviations between a given point and the centroid of a distribution. MD has been used for detecting multivariate outliers in different application areas such as psychology \citep{leys2018detecting}, archaeology \citep{papageorgiou2020ceramic}, and geotechnics \citep{zheng2021probabilistic}.    

We propose a MD constraint that restricts the MD between the solution $\boldsymbol{x}$ and the training input data to be less than or equal to a parameter $\gamma$. Let $\Gamma(\boldsymbol{x})$ be the MD measure mentioned above, which is calculated as:   
 \begin{align}
    \ \Gamma(\boldsymbol{x})\ =\ \sqrt{(\boldsymbol{x}-\boldsymbol\mu)^\intercal\textbf{C}^{-1}{(\boldsymbol{x}-\boldsymbol\mu)}},
\end{align}
where $\boldsymbol\mu$ and \textbf{C} denote the mean vector and the covariance matrix of the training inputs, respectively. We propose adding constraint  
\begin{align}\label{constraint:MD}
\Gamma(\boldsymbol{x}) \leq \gamma
\end{align}
as a way of modeling the trust region. We remark that for implementation purposes, the squared form of function $\Gamma$ can be directly inserted into the model and solved via off-the-shelf optimization  solvers such as Gurobi \citep{gurobi2020}.

Parameter $\gamma$ controls the level of conservativeness of the model. Low values for $\gamma$ result in a smaller feasible region in which the solution to the optimization problem must be close to the training input data. Larger values for $\gamma$ allow the outcome for the optimization model to be farther away from the training distribution. If the training input data follows a multivariate normal distribution, then $\Gamma^2(x)$ follows a chi-square distribution with $n$ degrees of freedom, where $n$ equals the number of dimensions of the training inputs \citep{johnson2014applied}. In practice, a point $\hat{\boldsymbol{x}}$ is commonly labeled as a multivariate outlier if $\Gamma^2(\hat{\boldsymbol{x}})$ exceeds a chi-square critical value with $n$ degrees of freedom at significance level $\alpha$ in $\{0.05, 0.01, 0.001\}$. As a result, if the training input data follows a multivariate normal distribution, then a natural value for $\gamma$ is given by $\gamma=\sqrt{\chi^2_{n, 1-\alpha}}$. We remark that in settings in which the training inputs do not follow a multivariate distribution, constraints based on MD can still be useful after tuning the value of $\gamma$.  


\subsubsection{K-Nearest Neighbors Constraints} \label{subsec:PMO-KNN model} \hfill \break
Our third class of constraints considers the distance between the solution of the optimization model and the $K$ closest points from the input data, denoted as its $K$-nearest neighbors. This is a straightforward way to ensure that the solution is less likely to be labeled as an outlier as the solution will be close to the input data. This type of measures have been widely used for classification and regression problems \citep{friedman2017elements}. Among the multiple measures of distance available, we propose a formulation that considers the $\ell^1$-norm distance metric, which computes the sum of the absolute difference of the individual components of the vectors. 
 
Consider a training data set $\mathcal{S}^x$ composed of $q$ observations $\boldsymbol{\hat{x}}^j \in \mathbb{R}^n$ for $j \in \mathcal{J}$ (where $\mathcal{J}=\{1,...,q\}$ is the index set of the observations). We propose a mixed-integer linear formulation that incorporates a constraint that limits the average distance between the solution of the optimization model and its $K$-nearest neighbors to be less than or equal to a parameter $\gamma$.   
For all $i \in \mathcal{I}$ and $j \in \mathcal{J}$, let auxiliary decision variables $d^-_{ij}$ and $d^+_{ij}$ capture the deviation between the $i^{th}$ component of solution vector $\boldsymbol{x}$ and the $i^{th}$ component of sample point $\boldsymbol{\hat{x}}^j$. Let auxiliary decision variables $w_j$ represent the $\ell^1$ distance distance between $\boldsymbol{x}$ and sample point $\boldsymbol{\hat{x}}^j$. We introduce auxiliary binary decision variables $\pi_{kj}$ that take a value of $1$ if sample point $\boldsymbol{\hat{x}}^j$ is one of the $k^{th}$ nearest neighbors of solution vector $\boldsymbol{x}$, and auxiliary variables $z_k$ to capture the $\ell^1$ distance between $\boldsymbol{x}$ and the $k^{th}$ nearest point in $\mathcal{S}$ (for example, $z_1$ captures the distance between $\boldsymbol{x}$ and the closest sample point, $z_2$ captures the distance between $\boldsymbol{x}$ and the second closest sample point, and so forth). Our proposed formulation of KNN constraints is given by: 
\begin{subequations}
\begin{align}
& x_i + d^+_{ij} - d^-_{ij} = \hat{x}_i^j  &\forall i \in \mathcal{I},\ j \in \mathcal{J}  \label{constr:PMO-KNN1} \\
& w_j = \sum_{i\in \mathcal{I}} d^+_{ij} + d^-_{ij}  &\forall j \in \mathcal{J} \label{constr:PMO-KNN2} \\
& z_k \ge w_j - M (1 - \pi_{kj})  &\forall j \in \mathcal{J},\ k \in \set{1,\dots,K} \label{constr:PMO-KNN3} \\
& \sum_{j\in \mathcal{J}} \pi_{kj} = k &\forall k \in \set{1,\dots,K} \label{constr:PMO-KNN4} \\
& \frac{1}{K} \sum_{k \in \set{1,\dots,K}} z_k \le \gamma \label{constr:PMO-KNN5} \\
& \pi_{kj} \in  \{0,1\} &\forall j \in \mathcal{J},\ k \in \set{1,\dots,K} \label{constr:vartypeK} \\
& d^+_{ij}, d^-_{ij} \geq 0 &\forall i \in \mathcal{I},\ j \in \mathcal{J} \label{constr:vartypeK2}\\
& w_j \geq 0 &\forall j \in \mathcal{J} \label{constr:vartypeK3}\\
& z_k \geq 0 &\forall k \in \set{1,\dots,K} \label{constr:vartypeK4}
\end{align}
\end{subequations}
where $M$ is a sufficiently large constant. Constraints~\eqref{constr:PMO-KNN1}--\eqref{constr:PMO-KNN2} define the $w$-variables, while constraints~\eqref{constr:PMO-KNN3} -\eqref{constr:PMO-KNN4} define the $z$-variables. Note that for $k = 1$, exactly one $\pi_{1j}$ variable equals 1 and so exactly one constraint (\ref{constr:PMO-KNN3}) is activated, ensuring that at optimality $z_1$ equals the distance between $\boldsymbol{x}$ and the closest sample point. For $k = 2$, then exactly two $\pi_{2j}$ variables take a value of 1, activating two constraints \eqref{constr:PMO-KNN3}. As a result, $z_2$ must be greater than or equal to two of the distances between $\boldsymbol{x}$ and the sample points, captured by the $w$-variables. At optimality, $z_2$ will be exactly equal to the distance between $\boldsymbol{x}$ and the second closest point in the sample. The model follows the same logic for the remaining values of $k$. Finally, constraint (\ref{constr:PMO-KNN5}) ensures that the average distance to the $K$-nearest neighbors is less than or equal to threshold $\gamma$.

It can be seen that the size of the PMO model with KNN constraints, both in number of decision variables and constraints, grows with the size of the training data (similar to constraints using convex hull of the dataset). Hence, for large data sets solving these models would be challenging for commercial mixed-integer programming solvers. As a heuristic approach, one may only consider the top $\beta\%$ of the training data when incorporating KNN constraints.

\section{Experiments} \label{sec:experiments}

In order to test the effectiveness of the various approaches, we evaluate the solution quality obtained by solving PMO on a variety of functions with different trust region definitions.  Specifically, we utilize standard global optimization functions for which the optimal solution is known to create synthetic datasets.  For each function, we randomly sample inputs from multivariate normal distributions to create a training set.  For each input in the training set, we evaluate the known function (with noise), and use that to train  models to predict the outcome. We can then solve a PMO with the various trust-region constraints.  

This allows for a controlled experimental evaluation of the efficacy of each trust region approach.  Because we know the optimal value and we can generate a data set for training predictive models, the gap between the solution value obtained via PMO and the optimal value can be calculated easily and compared across the various approaches.  Furthermore, we can also compare the best point in the training data with the values obtained through PMO to see if our proposed approaches improve upon the simple best-in-sample solution. Finally, we can also provide an analysis of the solution time to see which trust-region constraints provide the smallest computational burden. 

We present two sets of experimental results on two synthetic datasets based on seven global optimization benchmark functions. Given a synthetic dataset generated from a global optimization benchmark function $f\left(\boldsymbol{x}\right)$, we first estimate the relationship (which we denote by $F\left(\boldsymbol{x}\right)$ as in the definition of PMO) between the independent variables (i.e., the decision variables in the optimization problem) and the dependent variable using predictive modeling, then solve the PMO model, record its solution as $\boldsymbol{x}^*$, and calculate the true objective value at the solution $f\left(\boldsymbol{x}^*\right)$. We evaluate the quality of a solution by calculating the \emph{true outcome} of the solution (i.e., $f\left(\boldsymbol{x}^*\right)$), and the \emph{optimality error} at the optimal solution (calculated as $\Delta=|F\left(\boldsymbol{x}^*\right)-f\left(\boldsymbol{x}^*\right)|$). Note that we do not report percent optimality gaps because some of the functions have zero and/or negative evaluations at solutions.

All predictive models are trained in Python using the \textbf{scikit-learn} package \citep{scikit-learn}, and all optimization models are solved with Gurobi 9.0 \citep{gurobi2020} in Windows 10 on an Intel(R) Xeon(R) Gold 6130 CPU @ 2.10GHz processor with 32 GB RAM. 

\subsection{Data Generation} \label{subsec:data generation}
We consider seven benchmark functions:
\begin{enumerate}
    \item \textbf{Beale}: $f(\boldsymbol{x})=(1.5-x_1+x_1x_2)^2+(2.25-x_1+x_1x_2^2)^2+(2.625-x_1+x_1x_2^3)^2$. The global minimum is located at $\boldsymbol{x_{\text{min}}}=(3,\ 0.5),\ f(\boldsymbol{x_{\text{min}}})=0$.
    \item \textbf{Peaks}: $f(\boldsymbol{x})=3(1-x_1)^2e^{-x_1^2-(x_2+1)^2}-10(\frac{x_1}{5}-x_1^3-x_2^5)e^{-x_1^2-x_2^2}-\frac{1}{3}e^{-(x_1+1)^2-x_2^2}$. The global minimum is located at $\boldsymbol{x_{\text{min}}}=(0.23,\ -1.63),\ f(\boldsymbol{x_{\text{min}}})=-6.55$.
    \item \textbf{Griewank}: $f(\boldsymbol{x})=\sum_{i=1}^{4} \frac{x_i^2}{4000} - \prod_{i=1}^{4} \cos\left(\frac{x_i}{\sqrt{i+1}}\right) + 1$. The global minimum is located at $\boldsymbol{x_{\text{min}}}=(0,0,0,0),\ f(\boldsymbol{x_{\text{min}}})=0$.
    \item \textbf{Powell}: $f(\boldsymbol{x}) = (x_1+10x_2)^2+5(x_3-x_4)^2+(x_2-2x_3)^4+10(x_1-x_4)^4$. The global minimum is located at $\boldsymbol{x_{\text{min}}}=(0,0,0,0),\ f(\boldsymbol{x_{\text{min}}})=0$.
    \item \textbf{Quintic}: $f(\boldsymbol{x})=\sum_{i=1}^{5} |x_i^5-3x_i^4+4x_i^3+2x_i^2-10x_i-4|$. The global minimum is located at $\boldsymbol{x_{\text{min}}}=(-1,-1,-1,-1,-1)\ \text{or}\ (2,2,2,2,2),\ f(\boldsymbol{x_{\text{min}}})=0$.
    \item \textbf{Qing}: $f(\boldsymbol{x})=\sum_{i=1}^8 (x_i^2-i)^2$. The global minimum is located at $\boldsymbol{x_{\text{min}}}=(1,\sqrt{2},\sqrt{3},\sqrt{4},\sqrt{5},\sqrt{6},\sqrt{7},\\
    \sqrt{8}),\ f(\boldsymbol{x_{\text{min}}})=0$.
    \item \textbf{Rastrigin}: $f(\boldsymbol{x})=\sum_{i=1}^{10} [x_i^2-10 \cos(2 \pi x_i)]$. The global minimum is located at $\boldsymbol{x_{\text{min}}}=(0,0,0,0,0,0,0,\\
    0,0,0),\ f(\boldsymbol{x_{\text{min}}})=0$.
\end{enumerate}
\textbf{Beale}, \textbf{Griewank}, \textbf{Powell}, \textbf{Rastrigin}, \textbf{Qing}, and \textbf{Quintic} \citep{TestFunctinoIndex} are commonly used for evaluating global optimization algorithms, and \textbf{Peaks} is a test function in Matlab \citep{peaks}. 

For each function, we generate points to be used as training data for estimating the relationship $f\left(\boldsymbol{x}\right)$ by sampling $1000$ points from a multivariate normal distribution, with mean located at the global minimum point (i.e., $(3,0.5)$ for \textbf{Beale}, $(0.23,-1.63)$ for \textbf{Peaks}, $(0,0,0,0)$ for \textbf{Griewank}, $(0,0,0,0)$ for \textbf{Powell}, $(2,2,2,2,2)$ for \textbf{Quintic}, $(1,\sqrt{2},\sqrt{3},\sqrt{4},\sqrt{5},\sqrt{6},\sqrt{7},\sqrt{8})$ for \textbf{Qing}, and $(0,0,0,0,0,0,0,0,0,0)$ for \textbf{Rastrigin}) and a covariance matrix randomly generated using function \texttt{make\_spd\_matrix} in \textbf{scikit-learn}. The outcome of each sampled point is the evaluation of the benchmark function, plus randomly generated noise (i.e., $y\left(\boldsymbol{x}\right) = f\left(\boldsymbol{x}\right) + \epsilon$). For each point, the noise $\epsilon$ is drawn from a zero-mean normal random variable with variance given by a constant $ \sigma_{f\left(\boldsymbol{x}\right)}$, where $\sigma_{f\left(\boldsymbol{x}\right)}$ is the variance of the evaluations of the benchmark function on the sampled points.  

For each function, we generate $10$ different datasets of $1000$ points, each generated with a different randomly drawn covariance matrix. Therefore, for 7 global functions, we have 10 collections of input data, for a total of $7 \times 10 = 70$ datasets.  This creates the first set of data. 

For the second dataset, we follow the same procedure, but remove points surrounding the global optimal solution, to evaluate how well the PMO approaches can perform when the optimal value is far away from the training set.  In particular, we generate $10$ different datasets of $1500$ points, each generated with a different randomly drawn covariance matrix, but now removing the data within a central region.  This is described and evaluated in Section~\ref{sec:robust}.

\subsection{Predictive Modeling and Optimization Setup}
For each dataset, we scale the independent variables to $[0,1]$, standardize the dependent variable, and randomly split the dataset into a training set and a test set in the ratio of $7:3$. Each training set is used to fit three different predictive models: linear regression, random forest, and neural networks, to be used in PMO. 

For the configurations of the neurons in neural networks models, we consider a structure with two hidden layers. The number of neurons in each hidden layer ranges in $\{1,2,...,10\}$. For each dataset, we select the best neural networks model out of 500 candidates using the \texttt{GridSearchCV} function in \textbf{scikit-learn}. For random forest models, we create a hyper-parameters grid with the number of trees varying in $\{10,20,...,100\}$ and the maximum depth of the tree ranging in $\{1,2,...,10\}$. We try  $10 \times 10=100$ combinations of parameters and select the best model using \texttt{GridSearchCV} . For linear regression models, we simply fit one single model using \textbf{scikit-learn} for each dataset. 

The average out-of-sample performance, measured as $R^2$, of the various predictive models are presented in Table \ref{table: experiment1 r2}. As we can see, the performance of linear regression is typically very poor for these functions due to the degree of nonlinearity, whereas random forest models and neural networks models have much better performance.

\begin{table}[H]
\centering
\begin{threeparttable}[]
\caption{Predictive models performances ($R^2$) for the test set of synthetic datasets}
\label{table: experiment1 r2}
\begin{tabular}{lccccccc}
\toprule[1.5pt]\\[-0.5ex]
\textbf{Predictive Model} & \textbf{Beale} & \textbf{Peaks} & \textbf{Griewank} & \textbf{Powell} & \textbf{Quintic}  & \textbf{Qing}  & \textbf{Rastrigin} \\[0.5ex] \midrule \\[-0.5ex]
Linear Regression & 0.029 & 0.023 & -0.009 & -0.022 & 0.053 & 0.055 & -0.017 \\[-0.5ex]
Random Forest & 0.305 & 0.445 & 0.416 & 0.360 & 0.390 & 0.390 & 0.152 \\[-0.5ex]
Neural Networks & -0.298\tnote{\textdaggerdbl} & 0.435 & 0.396 & 0.393 & 0.376 & 0.374 & 0.071 \\[0.5ex] \bottomrule[1.5pt] 
\end{tabular}
\begin{tablenotes}
  \small
  \item[\textdaggerdbl] If a single outlier is removed,  the $R^2$ is elevated to 0.297.
\end{tablenotes}
\end{threeparttable}
\end{table}

\subsection{Optimization Results} \label{subsection: Optimization Results} 
Each predictive model fitted to each synthetic dataset is an instance. For each instance, we compare PMO with the following configurations:
\begin{itemize}
    \item \textbf{\textit{BASE}}: no constraints
    \item \textbf{\textit{IF}}: \hyperref[subsec:PMO-IF model]{IF} constraints
    \item \textbf{\textit{SVM}}: \hyperref[subsec:PMO-SVM model]{SVM} constraints solved directly using BARON
    \item $\textbf{\textit{SVM}}_{BC}$: \hyperref[subsec:PMO-SVM model]{SVM} constraints solved with our heuristic \hyperref[algo:SVM-BC]{branch-and-cut} algorithm
    \item \textbf{\textit{CH}}: \hyperref[subsec:PMO-CH model]{CH} constraints
    \item \textbf{\textit{PCA}}: \hyperref[subsec:PMO-PCA model]{PCA} constraints. We also run experiments with  \hyperref[subsec:PMO-PCA model]{PCA} as a penalty term, the same way as \cite{mistry2021mixed} propose. We omit these results as they are  very similar to the ones reported modeling PCA via constraints.
    \item \textbf{\textit{MD}}: \hyperref[subsec:PMO-MD model]{MD} constraints
    \item \textbf{\textit{KNN}}: \hyperref[subsec:PMO-KNN model]{KNN} constraints
\end{itemize}

For all the model formulations, as the input data is scaled between 0 and 1, the domain of the decision variables is $[\boldsymbol{0,1}]$. 
For \textbf{\textit{IF}}, the isolation forest is pre-trained in Python using the \textbf{scikit-learn} package with all the hyperparameters setting as default. We set the tree depth threshold at $d=5$ for datasets based on \textbf{Beale} and \textbf{Peaks}, and $d=6$ for datasets based on \textbf{Griewank}, \textbf{Powell}, \textbf{Quintic}, \textbf{Qing}, and \textbf{Rastrigin}. We pick this parameter by approximately halving the average depth of training points across the trees. For \textbf{\textit{SVM}}, we train the one-class SVM classifier using the \textbf{scikit-learn} package in Python. The hyperparameter $\gamma$ is set as default in the training step. For $\textbf{\textit{SVM}}_{BC}$, we set the number of pieces of the discretized ranges $m=10$. For \textbf{\textit{PCA}}, since the training datasets are not in high dimensions, we set the number of leading vectors $k=n-1$ ($n$ is the number of independent variables in the training datasets) to make sure that a subspace is defined and meanwhile it can explain the majority of the variance in the training data, and set $\gamma=0.00001$. For \textbf{\textit{MD}} we set significance level $\alpha$ at $0.05$, i.e., $\gamma = \sqrt{\chi^2_{n,0.95}}$, as the input data follows a multivariate normal distribution. For \textbf{\textit{KNN}}, we set $K=1$ and $\gamma = 0.1n$. In a tuning process, we set $\beta=10$ to only use the top 10\% of the data (based on their value of their dependent variable) for formulating KNN constraints. 

\subsubsection{Solution Outcome Quality}\label{subsubsection: Solution Outcome Quality} \hfill \break
We measure solution quality using two metrics: \emph{true outcome} $f\left(\boldsymbol{x}^*\right)$ and \emph{optimality error} $\Delta=|F\left(\boldsymbol{x}^*\right)-f\left(\boldsymbol{x}^*\right)|$, where $\boldsymbol{x}^*$ denotes the optimal solution obtained via PMO. Table~\ref{table: experiment1 overview} reports the average improvement of solution quality for different PMO variants as compared to the \textbf{\textit{BASE}} model over the 210 instances tested. Overall, \textbf{\textit{IF}} is the best performer both in terms of true outcome and optimality error, followed by \textbf{\textit{SVM}}.   

\begin{table}[H]
\centering
\begin{threeparttable}[]
\caption{Improvement of solutions from PMO variant models in comparison with $\boldsymbol{BASE}$}
\label{table: experiment1 overview}
\begin{tabular}{lcc}
\toprule[1.5pt]\\[-0.5ex]
 & $\dfrac{f\left(\boldsymbol{x}^*_{\boldsymbol{BASE}}\right)-f\left(\boldsymbol{x}^*_{variant}\right)}{f\left(\boldsymbol{x}^*_{\boldsymbol{BASE}}\right)-f(\boldsymbol{x_{\text{min}}})}$   & $\dfrac{\Delta_{\boldsymbol{BASE}}-\Delta_{variant}}{\Delta_{\boldsymbol{BASE}}}$  \\[3ex] \midrule \\[-0.5ex]
\textbf{\textit{IF}}      & 73.0\% & 81.6\% \\[0.0ex]
\textbf{\textit{SVM}}     & 62.1\% & 75.2\% \\[.0ex]
$\textbf{\textit{SVM}}_{BC}$ & 64.3\% & 75.1\% \\[.0ex]
\textbf{\textit{CH}}      & 43.9\% & 49.9\% \\[.0ex]
\textbf{\textit{PCA}}     & 21.3\% & 23.3\% \\[.0ex]
\textbf{\textit{MD}}      & 46.2\% & 52.4\% \\[.0ex]
\textbf{\textit{KNN}}     & 58.2\% & 62.5\% \\[0.5ex] 
\bottomrule[1.5pt] 
\end{tabular}
\end{threeparttable}
\end{table}

Table~\ref{table: experiment1} reports the average \emph{true outcome} $f\left(\boldsymbol{x}^*\right)$ and the average \emph{optimality error} $\Delta$ for the solutions obtained from the optimization models over all instances, grouped by benchmark function and predictive model. We also report the average true value $f\left(\boldsymbol{x}\right)$ for the point in the training data with the  smallest observed outcome $y\left(\boldsymbol{x}\right)$ in the training datasets, referred to as `Best Sample'. The best-performing model in each combination of benchmark function and predictive model are indicated in bold and underlined for \emph{true outcome} and \emph{optimality error}, respectively.

\setlength\rotheadsize{1cm}
\begin{table}[h]
\caption{Comparison $f\left(\boldsymbol{x}^*\right)$ from the optimization models for the benchmark functions using two metrics: average true outcome $f(\boldsymbol{x}^*)$ and average \emph{optimality error}
$\Delta=|F(\boldsymbol{x}^*)-f(\boldsymbol{x}^*)|$}
\label{table: experiment1}
\small
\begin{adjustbox}{max width=\textwidth}
\begin{threeparttable}
\begin{tabular}{clcccccccccccccccccccc}
\toprule[1.5pt]\\[0.5ex]
\multicolumn{1}{l}{} &  & \multicolumn{2}{c}{\textbf{Beale}} &  & \multicolumn{2}{c}{\textbf{Peaks}} &  & \multicolumn{2}{c}{\textbf{Griewank}} &  & \multicolumn{2}{c}{\textbf{Powell}} &  & \multicolumn{2}{c}{\textbf{Quintic}} &  & \multicolumn{2}{c}{\textbf{Qing}} &  & \multicolumn{2}{c}{\textbf{Rastrigin}} \\ [0.5ex]
\cline{3-4} \cline{6-7} \cline{9-10} \cline{12-13} \cline{15-16} \cline{18-19} \cline{21-22} \\ [-1ex]
\multicolumn{1}{l}{} &  & \multicolumn{1}{c}{$f(\boldsymbol{x}^*)$} & \multicolumn{1}{c}{$\Delta$} &  & \multicolumn{1}{c}{$f(\boldsymbol{x}^*)$} & \multicolumn{1}{c}{$\Delta$} &  & \multicolumn{1}{c}{$f(\boldsymbol{x}^*)$} & \multicolumn{1}{c}{$\Delta$} &  & \multicolumn{1}{c}{$f(\boldsymbol{x}^*)$} & \multicolumn{1}{c}{$\Delta$} &  & \multicolumn{1}{c}{$f(\boldsymbol{x}^*)$} & \multicolumn{1}{c}{$\Delta$} &  & \multicolumn{1}{c}{$f(\boldsymbol{x}^*)$} & \multicolumn{1}{c}{$\Delta$} &  & \multicolumn{1}{c}{$f(\boldsymbol{x}^*)$} & \multicolumn{1}{c}{$\Delta$} \\ [0.5ex]
\hline \\[-1.5ex]
\parbox[t]{2mm}{\multirow{8}{*}{\rotatebox[origin=c]{90}{Linear Regression}}} & \textbf{\textit{BASE}} & 3408 & 4980 &  & 0.00 & 2.98 &  & 1.04 & 0.68 &  & 30014 & 31967 &  & 3941 & 7804 &  & 1781 & 2931 &  & 260 & 210 \\
 & \textbf{\textit{IF}} & 8 & 293 &  & \textbf{-2.01} & 0.45 &  & 0.61 & \ul{0.02} &  & \textbf{197} & \ul{278} &  & \textbf{47} & \ul{224} &  & 94 & 135 &  & 107 & 6 \\
 & \textbf{\textit{SVM}} & 3 & 257 &  & -1.64 & \ul{0.02} &  & 0.76 & 0.17 &  & 605 & 621 &  & 99 & 1226 &  & 80 & 323 &  & 115 & 22 \\
 & $\textbf{\textit{SVM}}_{BC}$ & \textbf{3} & \ul{184} &  & -1.80 & 0.27 &  & 0.76 & 0.15 &  & 434 & 358 &  & 83 & 1029 &  & \textbf{75} & \ul{129} &  & 106 & 3 \\
 & \textbf{\textit{CH}} & 635 & 1940 &  & -0.07 & 2.30 &  & 1.01 & 0.42 &  & 4994 & 5002 &  & 2565 & 4411 &  & 1037 & 1376 &  & 119 & 17 \\
 & \textbf{\textit{PCA}} & 1683 & 3055 &  & 0.00 & 2.22 &  & 1.01 & 0.51 &  & 17540 & 18234 &  & 5234 & 7744 &  & 1699 & 2563 &  & 254 & 196 \\
 & \textbf{\textit{MD}} & 35 & 967 &  & -0.17 & 1.93 &  & 0.89 & 0.29 &  & 4579 & 4633 &  & 2686 & 4761 &  & 1135 & 1606 &  & 130 & 34 \\
 & \textbf{\textit{KNN}} & 36 & 978 &  & -0.69 & 1.25 &  & \textbf{0.54} & 0.07 &  & 680 & 653 &  & 584 & 1991 &  & 233 & 493 &  & \textbf{101} & \ul{1} \\ [0.5ex] \hline \\ [-1.5ex]
\parbox[t]{2mm}{\multirow{8}{*}{\rotatebox[origin=c]{90}{Random Forest}}} & \textbf{\textit{BASE}} & 101 & 598 &  & -6.03 & 0.76 &  & 0.44 & 0.23 &  & 689 & 1105 &  & 196 & 475 &  & 236 & 329 &  & 84 & 34 \\
 & \textbf{\textit{IF}} & \textbf{2} & \ul{52} &  & -6.16 & 0.91 &  & \textbf{0.09} & 0.17 &  & \textbf{30} & \ul{56} &  & \textbf{63} & \ul{34} &  & \textbf{51} & \ul{5} &  & \textbf{58} & 10 \\
 & \textbf{\textit{SVM}} & 11 & 155 &  & -6.03 & 0.76 &  & 0.19 & 0.06 &  & 206 & 366 &  & 76 & 108 &  & 179\tnote{\textdagger} & 138 &  & 77\tnote{\textdagger} & \ul{7} \\
 & $\textbf{\textit{SVM}}_{BC}$ & 24 & 168 &  & -6.02 & \ul{0.75} &  & 0.15 & 0.11 &  & 180 & 197 &  & 69 & 219 &  & 152 & 145 &  & 64 & 8 \\
 & \textbf{\textit{CH}} & 101 & 598 &  & -6.03 & 0.76 &  & 0.38 & 0.15 &  & 490 & 824 &  & 138 & 411 &  & 135 & 204 &  & 68 & 13 \\
 & \textbf{\textit{PCA}} & 63 & 191 &  & \textbf{-6.35} & 1.09 &  & 0.39 & 0.18 &  & 591 & 913 &  & 232 & 505 &  & 233 & 317 &  & 74 & 23 \\
 & \textbf{\textit{MD}} & 26 & 350 &  & -6.03 & 0.76 &  & 0.27 & \ul{0.04} &  & 451 & 816 &  & 171 & 448 &  & 191 & 270 &  & 74 & 23 \\
 & \textbf{\textit{KNN}} & 34 & 452 &  & -6.03 & 0.76 &  & 0.35 & 0.13 &  & 496 & 838 &  & 138 & 347 &  & 181 & 250 &  & 72 & 19 \\ [0.5ex] \hline \\ [-1.5ex]
\parbox[t]{2mm}{\multirow{8}{*}{\rotatebox[origin=c]{90}{Neural Networks}}} & \textbf{\textit{BASE}} & 2384 & 3050 &  & -6.30 & 0.79 &  & 0.82 & 1.10 &  & 9026 & 11487 &  & 5069 & 7451 &  & 2226 & 2589 &  & 238 & 166 \\
 & \textbf{\textit{IF}} & \textbf{9} & 5 &  & -6.32 & 0.46 &  & \textbf{0.13} & \ul{0.16} &  & \textbf{107} & \ul{284} &  & 81 & \ul{556} &  & \textbf{48} & \ul{74} &  & 121 & 27 \\
 & \textbf{\textit{SVM}} & 29 & 23 &  & -6.30 & 0.79 &  & 0.36 & 0.42 &  & 280 & 739 &  & \textbf{56} & 908 &  & 108\tnote{\textdagger} & 206 &  & 114\tnote{\textdagger} & 20 \\
 & $\textbf{\textit{SVM}}_{BC}$ & 13 & \ul{3} &  & -6.29 & 0.78 &  & 0.30 & 0.35 &  & 135 & 553 &  & 65 & 893 &  & 178\tnote{\textdagger} & 219 &  & 105\tnote{\textdagger} & \ul{8} \\
 & \textbf{\textit{CH}} & 654 & 1109 &  & -6.38 & 0.70 &  & 0.44 & 0.50 &  & 1015 & 1738 &  & 231 & 1255 &  & 132 & 216 &  & 102 & 10 \\
 & \textbf{\textit{PCA}} & 220 & 854 &  & \textbf{-6.44} & \ul{0.42} &  & 0.43 & 0.60 &  & 3717 & 5145 &  & 5723 & 7248 &  & 2019 & 2352 &  & 211 & 137 \\
 & \textbf{\textit{MD}}  & 56 & 395 &  & -6.30 & 0.79 &  & 0.49 & 0.55 &  & 1137 & 2036 &  & 401 & 1489 &  & 138 & 262 &  & 124 & 37 \\
 & \textbf{\textit{KNN}} & 25 & 365 &  & -6.36 & 0.73 &  & 0.16 & 0.19 &  & 624 & 1117 &  & 126 & 1066 &  & 101 & 173 &  & \textbf{85} & 9 \\ [0.5ex] \hline \\ [-1.5ex]
\multicolumn{2}{l}{Best Sample} & 2 &  &  & -4.95 &  &  & 0.22  &  &  & 110 & & & 236 & & & 54 & & & 63 & \\[1ex] \bottomrule[1.5pt]
\end{tabular}
\begin{tablenotes}
\small
\item[\textdagger] We are unable to obtain a feasible solution in 1, 4, 1, 3, 3, and 2 instances out of 10 instances within the time limit of 60 minutes for the combinations of \textbf{Qing} with random forest using \textbf{\textit{SVM}}, \textbf{Rastrigin} with random forest using \textbf{\textit{SVM}}, \textbf{Qing} with neural networks using \textbf{\textit{SVM}} and $\textbf{\textit{SVM}}_{BC}$, and \textbf{Rastrigin} with neural networks using \textbf{\textit{SVM}} and $\textbf{\textit{SVM}}_{BC}$, respectively.
\end{tablenotes}
\end{threeparttable}
\end{adjustbox}
\end{table}

We highlight several interesting observations. First and foremost, the quality of the solutions obtained through PMO is significantly better when trust-region constraints are added, independently of the predictive model chosen. Over the 210 instances tested, \textbf{\textit{BASE}} achieves solutions better than the seven PMO variant models in 15, 38, 26, 36, 74, 36, 30 instances and is worse in 194, 172, 177, 146, 95, 159, 173 instances, respectively, for \textbf{\textit{IF}}, \textbf{\textit{SVM}}, $\textbf{\textit{SVM}}_{BC}$, \textbf{\textit{CH}}, \textbf{\textit{PCA}}, \textbf{\textit{MD}}, and \textbf{\textit{KNN}}. From Table~\ref{table: experiment1} we also see that \textbf{\textit{BASE}} never achieves the best solution quality for any benchmark function and predictive model combination. This makes a clear case for why incorporating trust-region constraints is necessary in the context of PMO.  

Furthermore, PMO variant models are more likely to identify solutions that are better than the best points in the training data, which suggests that the generalization of the predictive models is enhanced when solutions are constrained to a trust region. As an interesting example, for \textbf{Quintic} using linear regression, the average true outcome of the solutions obtained from \textbf{\textit{BASE}} is 3,941, which is far away from the global minimum of 0. In comparison, \textbf{\textit{MD}} improves the average true outcome to 2,686, \textbf{\textit{KNN}} enhances the value to 584, \textbf{\textit{SVM}} further pushes it to 99, and \textbf{\textit{IF}} finds solutions closer to the global minimum with the average true outcome of 47.

 \textbf{\textit{IF}} delivers the most robust performance in terms of \emph{true outcome} among all the models; it identifies the highest quality solution in average in 13 out of 21 combinations of benchmark function and predictive model. For the remaining eight scenarios, the performance of \textbf{\textit{IF}} is marginally inferior to the best model. One can also infer from Table~\ref{table: experiment1 overview} and Table~\ref{table: experiment1} that $\textbf{\textit{SVM}}_{BC}$ is  the second best model tested.
\begin{figure}[h]
    \subfigure[Linear Regression]{
        \includegraphics[width=.31\textwidth]{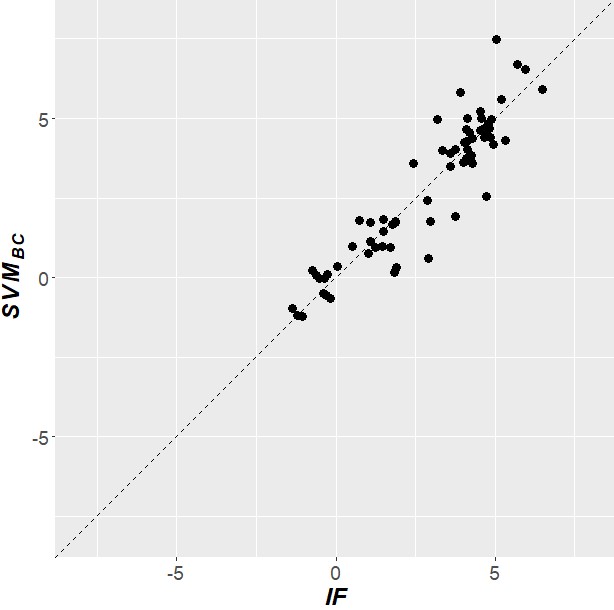}
        \label{fig:scatterplot_LR}}
    \subfigure[Random Forest]{
        \includegraphics[width=.31\textwidth]{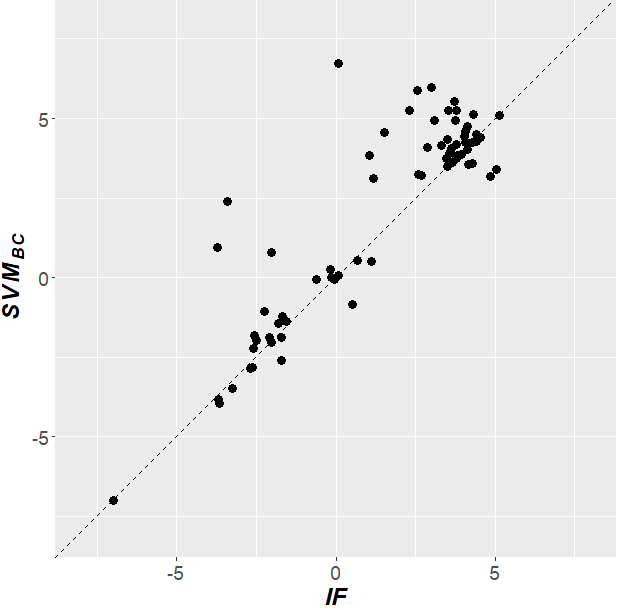}        \label{fig:scatterplot_RF}}
    \subfigure[Neural Networks]{
        \includegraphics[width=.31\textwidth]{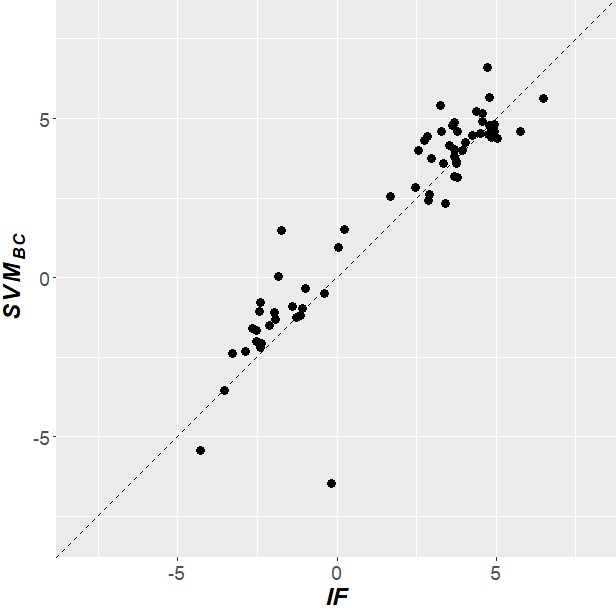}
        \label{fig:scatterplot_NN}}
    \caption{Scatter plots comparing the gap from true outcome to global minimum, $f\left(\boldsymbol{x}^*\right) - f\left(\boldsymbol{x_{\text{min}}}\right)$, between $\boldsymbol{IF}$ and $\boldsymbol{SVM}_{BC}$ base on different predictive models in natural logarithmic scale}
    \label{fig:scatterplot of trueoutcome}
\end{figure}
To compare the two best models, a detailed analysis between \textbf{\textit{IF}} and $\textbf{\textit{SVM}}_{BC}$ grouped by the predictive models used in PMO is provided in Figure~\ref{fig:scatterplot of trueoutcome}, in which three scatter plots comparing \emph{true optimality gap} of solutions obtained from \textbf{\textit{IF}} and $\textbf{\textit{SVM}}_{BC}$ (calculated as log($f\left(\boldsymbol{x}^*\right)-f\left(\boldsymbol{x_{\text{min}}}\right)$)) are provided. Each point corresponds to an instance. A point above the dashed line at 45 degrees indicates that the true outcome of the solution obtained from \textbf{\textit{IF}} is better than that from $\textbf{\textit{SVM}}_{BC}$. When the predictive model is linear regression (Figure~\ref{fig:scatterplot_LR}), random forest (Figure~\ref{fig:scatterplot_RF}), and neural networks (Figure \ref{fig:scatterplot_NN}), \textbf{\textit{IF}} is better than $\textbf{\textit{SVM}}_{BC}$ in 36, 45, and 49 out of 70 instances, respectively. 

Of note is that in Figure~\ref{fig:scatterplot_LR} we see that \textbf{\textit{IF}} and $\textbf{\textit{SVM}}_{BC}$ have a similar performance when linear regression is employed as the predictive model. We hypothesize that the reason for such similar solution quality is the poor performance of linear regression in this setting (see Table~\ref{table: experiment1 r2}). The performance difference between \textbf{\textit{IF}} and $\textbf{\textit{SVM}}_{BC}$ is more obvious when the predictive model has higher performance. Figure \ref{fig:scatterplot_RF} shows that with random forest as the predictive model the points tend to the bottom left corner and so are of better quality on average, but that the points deviate more from the 45 degrees dashed line in the vertical direction than in the horizontal direction, meaning that the \emph{true optimality gap} from \textbf{\textit{IF}} is much better than that from $\textbf{\textit{SVM}}_{BC}$. An analogous pattern can be observed in Figure~\ref{fig:scatterplot_NN}. Overall, Figure \ref{fig:scatterplot of trueoutcome} clearly indicates the superiority of \textbf{\textit{IF}} over the other tested models on the synthetic datasets. In addition, such dominance of \textbf{\textit{IF}} can also be found in terms of the optimality error, which measures the level of overestimation at the solutions \citep{schweidtmann2022obey} and can be interpreted as the reliability of the solutions. According to Table~\ref{table: experiment1}, the performance of  \textbf{\textit{IF}} in terms of the optimality error is the most robust among all the models; it is the best among those models tested in 11 out of 21 scenarios. The overall improvement as compared to \textbf{\textit{BASE}}, which is summarized in Table~\ref{table: experiment1 overview}, also shows that \textbf{\textit{IF}} is the best of all the PMO variant models.

A final takeaway is that \textbf{\textit{MD}} and \textbf{\textit{KNN}} are two alternative models worth exploration for any particular application. Although the performance of \textbf{\textit{MD}} and \textbf{\textit{KNN}} are unable to surpass \textbf{\textit{SVM}} in terms of \emph{true outcome}, both of them are better than \textbf{\textit{PCA}} in 18 cases out of 21, and are better than  \textbf{\textit{CH}} in 9 cases and 16 cases out of 21, respectively. In addition, as summarized in Table~\ref{table: experiment1 overview}, \textbf{\textit{KNN}} ranks fourth place among the PMO variant models, only slightly worse than the best model in literature. Such good performance can also be supported by Table~\ref{table: experiment1}, through the results from \textbf{Rastrigin} in particular, which has 10 dimensions with the predictive model being linear regression and neural networks.
 


 

\subsubsection{Computational Efficiency} \hfill \break
From the results of solution outcome values reported in the Section \ref{subsubsection: Solution Outcome Quality}, we observe that solutions from \textbf{\textit{IF}}, \textbf{\textit{SVM}}, and $\textbf{\textit{SVM}}_{BC}$ are significantly better than other PMO variant models, with \textbf{\textit{IF}} being the best approach in terms of solution outcome quality. 

\begin{figure}[H]
	\centering
	\captionsetup{width=\linewidth}
	\includegraphics[scale=0.6]{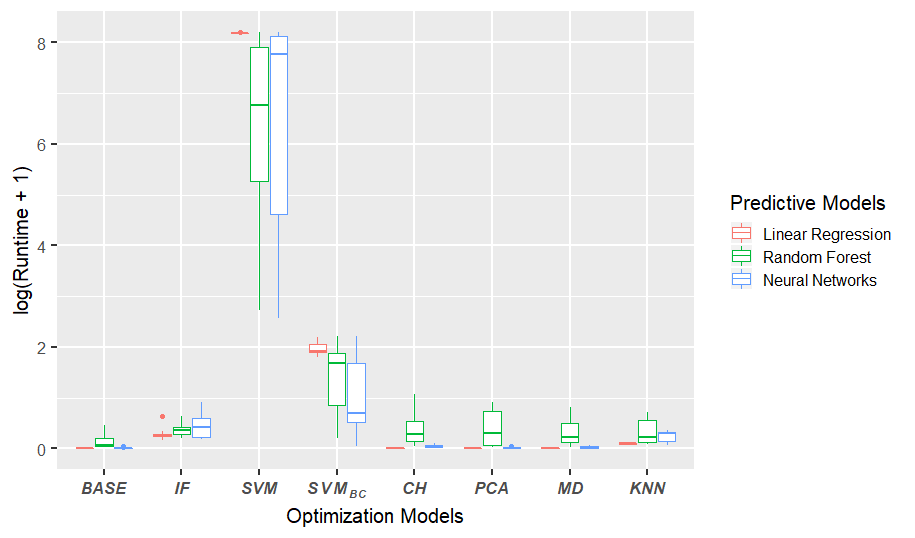}
	\caption{Solution times of instances generated with \textbf{Powell} and three predictive models using different optimization models in natural logarithmic scale}
	\label{fig:runtime_Powell}
\end{figure}

Now we discuss the solution times for all models. Figure~\ref{fig:runtime_Powell} shows a box plot of running time over all 10 instances using \textbf{Powell} as the bechmark function, with the three tested predictive models and the seven PMO variant models. Overall, Figure~\ref{fig:runtime_Powell} highlights the computational efficiency of our \textbf{\textit{IF}} model over the other models that achieve high-quality solutions based on SVMs.  The running time of the majority of the instances using \textbf{\textit{MD}}, \textbf{\textit{CH}}, \textbf{\textit{KNN}}, and \textbf{\textit{PCA}}, and \textbf{\textit{IF}} are within two seconds, but as discussed in the previous section, \textbf{\textit{IF}} finds much better solutions. In contrast, the run time for \textbf{\textit{SVM}} is significantly longer than other PMO variant models. For example, when the predictive model is linear regression, \textbf{\textit{SVM}} uses nearly the full 60-minute time limit for all the 10 instances , among which only 2 instances are solved to optimality. Our implementation of $\textbf{\textit{SVM}}_{BC}$ improves the solution time of \textbf{\textit{SVM}} significantly, but it still lags \textbf{\textit{IF}} even for \textbf{Powell} which only has 4 dimensions. 
\begin{figure}[H]
	\centering
	\captionsetup{width=\linewidth}
	\includegraphics[scale=0.58]{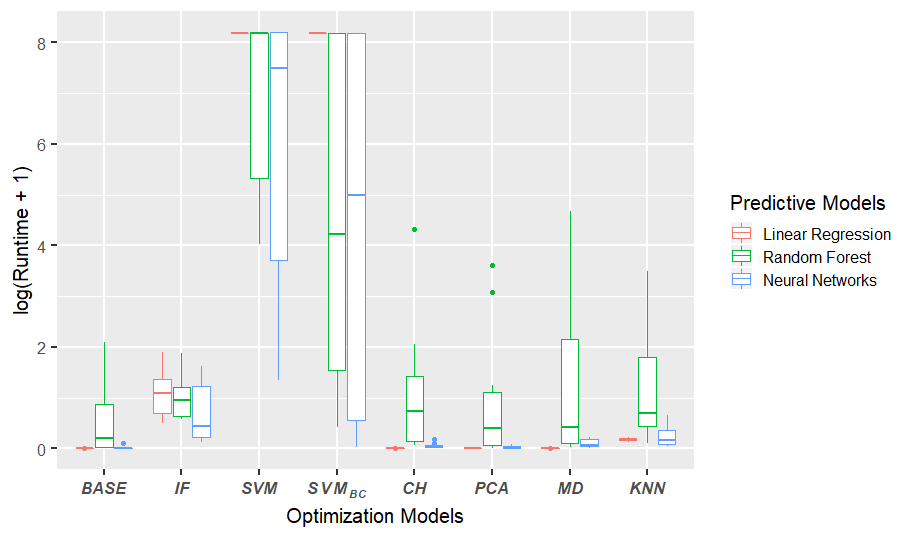}
	\caption{Solution times of instances generated with \textbf{Qing} and three predictive models using different optimization models in natural logarithmic scale}
	\label{fig:runtime_Qing}
\end{figure}
As expected, for the functions with more dimensions, the run time performance degrades. Figure \ref{fig:runtime_Qing} summarizes the runtimes for \textbf{Qing} function with 8 dimensions. Despite the overall increase in run time, we see that except \textbf{\textit{SVM}} and $\textbf{\textit{SVM}}_{BC}$, the \textbf{Qing} instances are solved within the time limit by all the remaining models, among which, \textit{\textbf{IF}} exhibits a robust performance regardless of the predictive model. Appendix~\ref{sec:appendix-running time} summarizes the run time performance over all benchmark functions, which particularly highlight the robust computational performance of \textit{\textbf{IF}}. We should note that, for \textbf{Rastrigin} function, computational times across all constraint types for the random forest model are higher. This is mainly due to the instance complexities for the random forest-based PMO model even without any trust-region constraint. 

\begin{figure}[h]
	\centering
	\captionsetup{width=\linewidth}
	\includegraphics[scale=0.50]{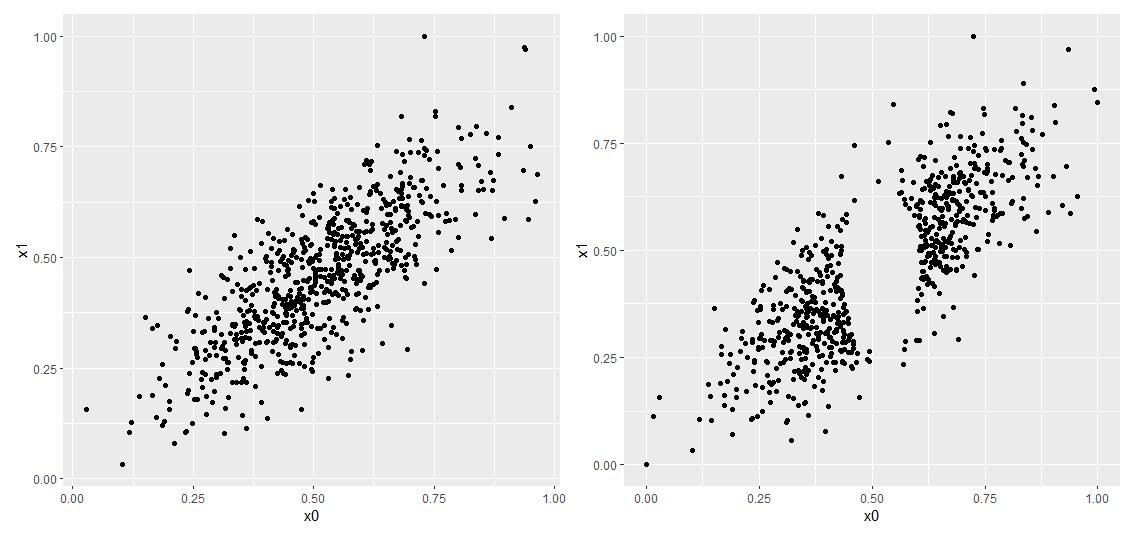}
	\caption{Comparison of shapes of datasets simulated with \textbf{Peaks} function using same parameter setting, with the left figure depicting the first experimental setting and the right figure showing the result of removing points in the optimum.}
	\label{fig:illustrative_data_comparison}
\end{figure}
\subsection{Robustness Check: Sampling Away from Global Optimum} \label{sec:robust}
In addition to the experimental results from the 70 synthetic datasets, we also run a similar experiment on another 70 synthetic datasets by taking a different random sampling approach. To be more specific, in these datasets we exclude the samples in the neighborhood of the global optimum. 
With the same seven benchmark functions, we first generate 1500 sample points instead of 1000 sample points for each dataset using the same sampling procedure in Section~\ref{subsec:data generation}. We then create a hole centering at the global optimum of the benchmark function 
with radius $r = \frac{1}{2} \sqrt{\sigma_1^2+...+\sigma_n^2}$, where $\sigma_i^2$ is the variance of independent variable $x_i$ and is obtained from the randomly generated covariance matrix. Finally, we remove sample points falling within the radius. Figure \ref{fig:illustrative_data_comparison} illustrates the shapes of two datasets simulated using \textbf{Peaks} function.

\setlength\rotheadsize{1cm}
\begin{table}[H]
\caption{Comparison $f\left(\boldsymbol{x}^*\right)$ from the optimization models for the benchmark functions using two metrics: average true outcome $f(\boldsymbol{x}^*)$ and average \emph{optimality error}
$\Delta=|F(\boldsymbol{x}^*)-f(\boldsymbol{x}^*)|$}
\label{table: experiment2}
\small
\begin{adjustbox}{max width=\textwidth}
\begin{threeparttable}
\begin{tabular}{clcccccccccccccccccccc}
\toprule[1.5pt]\\[0.5ex]
\multicolumn{1}{l}{} &  & \multicolumn{2}{c}{\textbf{Beale}} &  & \multicolumn{2}{c}{\textbf{Peaks}} &  & \multicolumn{2}{c}{\textbf{Griewank}} &  & \multicolumn{2}{c}{\textbf{Powell}} &  & \multicolumn{2}{c}{\textbf{Quintic}} &  & \multicolumn{2}{c}{\textbf{Qing}} &  & \multicolumn{2}{c}{\textbf{Rastrigin}} \\ [0.5ex]
\cline{3-4} \cline{6-7} \cline{9-10} \cline{12-13} \cline{15-16} \cline{18-19} \cline{21-22} \\ [-1ex]
\multicolumn{1}{l}{} &  & \multicolumn{1}{c}{$f(\boldsymbol{x}^*)$} & \multicolumn{1}{c}{$\Delta$} &  & \multicolumn{1}{c}{$f(\boldsymbol{x}^*)$} & \multicolumn{1}{c}{$\Delta$} &  & \multicolumn{1}{c}{$f(\boldsymbol{x}^*)$} & \multicolumn{1}{c}{$\Delta$} &  & \multicolumn{1}{c}{$f(\boldsymbol{x}^*)$} & \multicolumn{1}{c}{$\Delta$} &  & \multicolumn{1}{c}{$f(\boldsymbol{x}^*)$} & \multicolumn{1}{c}{$\Delta$} &  & \multicolumn{1}{c}{$f(\boldsymbol{x}^*)$} & \multicolumn{1}{c}{$\Delta$} &  & \multicolumn{1}{c}{$f(\boldsymbol{x}^*)$} & \multicolumn{1}{c}{$\Delta$} \\ [0.5ex]
\hline \\[-1.5ex]
\parbox[t]{2mm}{\multirow{8}{*}{\rotatebox[origin=c]{90}{Linear Regression}}} & \textbf{\textit{BASE}} & 9375 & 11320 &  & 0.00 & 2.16 &  & 1.03 & 0.48 &  & 16121 & 17009 &  & 3835 & 7623 &  & 1992 & 3507 &  & 278 & 223 \\
 & \textbf{\textit{IF}} & 11 & 367 &  & -0.39 & 0.14 &  & \textbf{0.61} & \ul{0.18} &  & \textbf{194} & 584 &  & \textbf{68} & \ul{411} &  & 81 & 151 &  & \textbf{108} & \ul{3} \\
 & \textbf{\textit{SVM}} & 8 & 444 &  & \textbf{-0.85} & \ul{0.10} &  & 0.96 & 0.23 &  & 667 & 251 &  & 116 & 1190 &  & 118 & 578 &  & 118 & 22 \\
 & $\textbf{\textit{SVM}}_{BC}$ & \textbf{4} & \ul{172} &  & -0.72 & 0.15 &  & 0.90 & 0.17 &  & 603 & \ul{113} &  & 102 & 1044 &  & \textbf{61} & \ul{128} &  & 113 & 9 \\
 & \textbf{\textit{CH}} & 581 & 2170 &  & -0.02 & 1.35 &  & 0.92 & 0.20 &  & 7230 & 7074 &  & 3005 & 5008 &  & 1045 & 1415 &  & 142 & 36 \\
 & \textbf{\textit{PCA}} & 1546 & 3184 &  & 0.00 & 1.07 &  & 1.04 & 0.42 &  & 15585 & 15975 &  & 5369 & 8206 &  & 2301 & 3370 &  & 267 & 194 \\
 & \textbf{\textit{MD}} & 66 & 1383 &  & -0.03 & 1.13 &  & 0.93 & 0.21 &  & 13355 & 13278 &  & 3770 & 6024 &  & 1260 & 1756 &  & 126 & 24 \\
 & \textbf{\textit{KNN}} & 36 & 1434 &  & -0.25 & 0.94 &  & 0.74 & \ul{0.01} &  & 1575 & 1197 &  & 1552 & 3095 &  & 265 & 586 &  & 110 & 6 \\ [0.5ex] \hline \\ [-1.5ex]
\parbox[t]{2mm}{\multirow{8}{*}{\rotatebox[origin=c]{90}{Random Forest}}} & \textbf{\textit{BASE}} & 53 & 627 &  & -2.39 & 0.04 &  & 0.48 & 0.23 &  & 502 & 1069 &  & 250 & 326 &  & 254 & 255 &  & 84 & 25 \\
 & \textbf{\textit{IF}} & \textbf{15} & \ul{38} &  & -1.92 & 0.10 &  & \textbf{0.23} & 0.18 &  & \textbf{115} & \ul{125} &  & \textbf{95} & 48 &  & \textbf{54} & \ul{7} &  & 60 & 21 \\
 & \textbf{\textit{SVM}} & 80 & 152 &  & -2.29 & 0.29 &  & 0.30 & \ul{0.10} &  & 219 & 450 &  & 113 & \ul{13} &  & 109\tnote{\textsection} & 40 &  & 91\tnote{\textsection} & 17 \\
 & $\textbf{\textit{SVM}}_{BC}$ & 16 & 83 &  & -2.63 & 0.64 &  & 0.26 & 0.14 &  & 198 & 377 &  & 119 & 16 &  & 90 & 72 &  & \textbf{55} & 12 \\
 & \textbf{\textit{CH}} & 27 & 558 &  & -2.42 & \ul{0.02} &  & 0.46 & 0.20 &  & 577 & 1085 &  & 118 & 123 &  & 157 & 148 &  & 67 & \ul{1} \\
 & \textbf{\textit{PCA}} & 265 & 643 &  & \textbf{-4.57} & 2.55 &  & 0.54 & 0.26 &  & 842 & 1313 &  & 285 & 355 &  & 215 & 211 &  & 84 & 24 \\
 & \textbf{\textit{MD}} & 154 & 553 &  & -2.48 & 0.16 &  & 0.40 & {\ul 0.12} &  & 750 & 1261 &  & 232 & 252 &  & 163 & 158 &  & 75 & 16 \\
 & \textbf{\textit{KNN}} & 39 & 536 &  & -2.24 & 0.20 &  & 0.44 & 0.18 &  & 458 & 944 &  & 152 & 128 &  & 147 & 131 &  & 65 & 3 \\ [0.5ex] \hline \\ [-1.5ex]
\parbox[t]{2mm}{\multirow{8}{*}{\rotatebox[origin=c]{90}{Neural Networks}}} & \textbf{\textit{BASE}} & 8713 & 9179 &  & -0.99 & 5.15 &  & 1.03 & 1.47 &  & 5059 & 6742 &  & 3789 & 5714 &  & 2273 & 2802 &  & 262 & 190 \\
 & \textbf{\textit{IF}} & \textbf{5} & 104 &  & -3.76 & 0.59 &  & \textbf{0.18} & \ul{0.03} &  & \textbf{162} & \ul{291} &  & \textbf{148} & \ul{204} &  & \textbf{63} & \ul{155} &  & 111 & 12 \\
 & \textbf{\textit{SVM}} & 9 & 133 &  & -3.55 & \ul{0.34} &  & 0.47 & 0.37 &  & 409 & 792 &  & 309 & 483 &  & 92 & 274 &  & 107\tnote{\textsection} & 7 \\
 & $\textbf{\textit{SVM}}_{BC}$ & 12 & \ul{89} &  & -3.66 & 0.47 &  & 0.38 & 0.27 &  & 324 & 653 &  & 187 & 504 &  & 73 & 226 &  & 101 & \ul{1} \\
 & \textbf{\textit{CH}} & 406 & 703 &  & -2.53 & 1.35 &  & 0.40 & 0.36 &  & 789 & 1320 &  & 216 & 814 &  & 222 & 352 &  & 106 & 9 \\
 & \textbf{\textit{PCA}} & 1038 & 1214 &  & \textbf{-3.88} & 0.65 &  & 0.94 & 1.04 &  & 2708 & 4094 &  & 2547 & 3851 &  & 1285 & 1658 &  & 228 & 151 \\
 & \textbf{\textit{MD}}  & 138 & 387 &  & -2.99 & 0.55 &  & 0.51 & 0.50 &  & 911 & 1374 &  & 267 & 1012 &  & 120 & 291 &  & 103 & 12 \\
 & \textbf{\textit{KNN}} & 157 & 407 &  & -2.36 & 0.84 &  & 0.30 & 0.27 &  & 358 & 870 &  & 170 & 652 &  & 144 & 263 &  & \textbf{100} & 6  \\ [0.5ex] \hline \\ [-1.5ex]
\multicolumn{2}{l}{Best Sample} & 34 &  &  & -1.91 &  &  & 0.31 &  &  & 334 &  &  & 283 &  &  & 158 &  &  & 69 & \\[1ex] \bottomrule[1.5pt]
\end{tabular}
\begin{tablenotes}
\small
\item[\textsection] We are unable to obtain a feasible solution in 1, 7, and 3 instances out of 10 within the time limit of 60 minutes for the combinations of \textbf{Qing} with random forest using \textbf{\textit{SVM}}, \textbf{Rastrigin} with random forest using $\textbf{\textit{SVM}}_{BC}$, and \textbf{Rastrigin} with neural networks using $\textbf{\textit{SVM}}_{BC}$, respectively.
\end{tablenotes}
\end{threeparttable}
\end{adjustbox}
\end{table}

Similar to the pattern that we observe in Table \ref{table: experiment1}, Table \ref{table: experiment2} exhibits the dominance of \textbf{\textit{IF}} among all the PMO variant models; it provides the best solution in terms of \emph{true outcome} in 14 cases out of 21. \textbf{\textit{IF}} also is the best in terms of \emph{optimality error}.

\section{Conclusions and Future Work} \label{sec:conclusion}
In this paper, we investigate seven PMO variant models with constraints describing a trust region. We assess the performance of both the constraints existing in literature and our proposed constraints, validate the importance of considering a trust region, and highlight that constraints learned from isolation forests outperform existing approaches in both solution quality and computational performance through extensive experimentation on several synthetic datasets.

Among the existing approaches, one-class SVM constraints work best in terms of solution quality but impose a high computational burden to the PMO models. In addition to isolation forest, we explore two other common outlier detection methods, Mahalanobis distance and $K$-nearest neighbors. The performance of trust-region constraints based on these two methods is better than two other existing approaches, namely, convex hull constraints and PCA constraints.

There is a wide array of research that can expand on the work in this paper, including the following. First, we do not investigate specific algorithms well suited for each predictive model and each PMO variant model. We observed that \textbf{\textit{SVM}} suffers from the complexity brought by the exponential constraints. Trust-region constraints in other PMO variant models could also potentially significantly increase the solution times (e.g., consider the auxiliary binary variables and big-M constraints for \textbf{\textit{KNN}}). Advanced optimization algorithms and stronger formulations may be applicable to greatly improve computational times. Second, an investigation of which type of trust-region constraint to add for different contexts is important. We show that adding trust-region constraints nearly always leads to higher quality solutions in our experiments, but using different constraints results in varied solution quality and understanding this connection would make the decision-making pipeline more robust. Third, we explored three classes of trust-region constraints based on existing outlier detection methods. There are a variety of additional methods from the literature of outlier detection, which can be used for characterizing a trust region and could offer different tradeoffs between solution quality and computational performance.


\clearpage

\begin{appendices} {}
\section{PMO Formulations with Different Predictive Models} \label{sec:appendix1}
\subsection{Linear Regression-based PMO Formulation}
For a linear regression model, the objective function can be expressed as $F(\boldsymbol{x}) = {b_0+b_1x_i+...+b_nx_n}$. As a result, the PMO formulation is given by the following linear program:
\begin{subequations}
\begin{align}
\min_{\boldsymbol{x}}  \quad & b_0+b_1x_1+...+b_nx_n \tag{A.1a} \label{A.1a}\\
\textrm{s.t.} \quad & \boldsymbol{x} \in \mathbf{[0,1]}. \tag{A.1b} \label{A.1b}
\end{align}
\end{subequations}

\subsection{Random Forest-based PMO Formulation}

\subsubsection{Background on Random Forests} \hfill \break
We consider random forest regression models in this paper, where the outcome variable is continuous. Let $\mathcal{T}^{\text{RF}}$ be the index set of trees in the random forest. Each tree $t\in \mathcal{T}^{\text{RF}}$ has the same weight in the prediction and is trained over a randomly selected subsample of the training data. Similar to isolation trees, the prediction for an observation is determined by checking a series of splits. Each leaf node $\ell$ from tree $t \in \mathcal{T}^{\text{RF}}$ has a score $\mathsf{p}_{t,\ell}$. Any vector of independent variables $\boldsymbol{x}$ corresponds to a leaf node of the tree, denoted by $\ell(\boldsymbol{x})$. The predicted outcome for an input vector $\boldsymbol{x}$ from tree $t$ is given by the corresponding score $\mathsf{p}_{t,\ell(\boldsymbol{x})}$. As a result, the prediction of the random forest for a given $\boldsymbol{x}$ is given by the average of the corresponding scores: $$F(\boldsymbol{x}) = \frac{1}{|\mathcal{T}^{\text{RF}}|}  \sum_{t \in \mathcal{T}^{\text{RF}}}\mathsf{p}_{t,\ell(\boldsymbol{x})}$$

\subsubsection{Optimization Model} \hfill \break
Inspired by \cite{mivsic2020optimization}, we present a PMO formulation to solve optimization problems over a pre-trained random forest. The objective of the optimization problem is to find the independent variable vector $\boldsymbol{x}$ that minimizes the outcome predicted by the random forest $F(\boldsymbol{x})$. This can be casted as the problem of finding a leaf node from each tree in the random forest that minimize the average outcome.

Consider a random forest with $|\mathcal{T}^{\text{RF}}|$ trees. Let $\mathcal{L}^{\text{RF}}_t$ be the set of leaf nodes in tree $t \in \mathcal{T}^{\text{RF}}$. We introduce binary decision variable $\mathsf{y}_{t,\ell}$ that takes a value of 1 if leaf node $\ell$ from tree $t \in \mathcal{T}^{\text{RF}}$ is selected. We introduce auxiliary decision variables ($\mathsf{z}_i^{LB},\ \mathsf{z}_i^{UB}$) to specify the range of each variable feature $i \in \mathcal{I}$. Let the lower and upper bounds corresponding to a leaf node be $l_{i,t,\ell}^{\text{RF}}$ and $u_{i,t,\ell}^{\text{RF}}$ for all $i \in \mathcal{I},\ t \in \mathcal{T}^{\text{RF}},\ \text{and } \ell \in \mathcal{L}^{\text{RF}}_t$. Our formulation of the random forest-based optimization problem is:
\begin{subequations}
\begin{align}
\min_{\boldsymbol{x, \mathsf{y},\mathsf{z}}}  \quad & \frac{1}{|\mathcal{T}^{\text{RF}}|} \sum_{t \in \mathcal{T}^{\text{RF}}}\sum_{\ell \in \mathcal{L}^{\text{RF}}_t}{\mathsf{p}_{t,\ell} \mathsf{y}_{t,\ell}} \tag{A.2a} \label{A.2a}\\
\textrm{s.t.} \quad & \sum_{\ell \in \mathcal{L}^{\text{RF}}_t}\mathsf{y}_{t,\ell}=1 &\forall t \in \mathcal{T}^{\text{RF}} \tag{A.2b} \label{A.2b}\\
& \sum_{\ell \in \mathcal{L}^{\text{RF}}_t}l_{i,t,\ell}^{\text{RF}} \: \mathsf{y}_{t,\ell} \le \mathsf{z}_i^{LB} &\forall t \in \mathcal{T}^{\text{RF}},\ \forall i \in \mathcal{I}\tag{A.2c} \label{A.2c}\\
& 1-\sum_{\ell \in \mathcal{L}^{\text{RF}}_t}(1-u_{i,t,\ell}^{\text{RF}}) \: \mathsf{y}_{t,\ell} \ge \mathsf{z}_i^{UB} &\forall t \in \mathcal{T}^{\text{RF}},\ \forall i \in \mathcal{I} \tag{A.2d}\label{A.2d}\\
& \mathsf{z}_i^{LB} \le x_i \le \mathsf{z}_i^{UB}- \epsilon \qquad &\forall i \in \mathcal{I} \tag{A.2e}\label{A.2e}\\
& x_i, \mathsf{z}_i^{LB}, \mathsf{z}_i^{UB} \in [0,1] &\forall i \in \mathcal{I} \tag{A.2f}\label{A.2f}\\
& \mathsf{y}_{t,\ell} \in \{0,1\} & \forall t \in \mathcal{T}^{\text{RF}},\ \ell \in \mathcal{L}^{\text{RF}}_t \tag{A.2g}\label{A.2g}
\end{align}
\end{subequations}

Constraints (\ref{A.2b}) ensure that exactly one of the leaf nodes from each tree in the random forest is selected. Constraints (\ref{A.2c}) and constraints (\ref{A.2d}) ensure that, for any leaf node that is selected, the domain of the solution is limited to the feature ranges determined by the path leading to that leaf node. Constraints (\ref{A.2e}) ensure that $\boldsymbol{x}$ belongs to the domain defined by the corresponding leaf nodes selected. 

\subsection{Neural Networks-based PMO Formulation}
There is plenty of literature on formulating neural networks models as a mixed integer programming problem. In particular, we make use of recent work that solve the neural networks-based optimization problem. For the details of the formulation, the reader is refered to \cite{bergman2022janos}. 

\section{Details of the Branch-and-Cut Algorithm for PMO Model with One-Class SVM Constraints} \label{appendix:appendix-svm-bc model}
Recall that $\boldsymbol{x} = \{x_i\}_{i=1}^n$ are the main decision variables in a PMO model, determining the value of each feature $i \in \mathcal{I}$. For the branch-and-cut algorithm, we  first discretize the range of each feature $i\in \mathcal{I}$ into \textit{m} pieces. As each feature is scaled to [0,1], the $m$ pieces of the discretized ranges for each feature are $[0, \frac{1}{m}], [\frac{1}{m}+\epsilon, \frac{2}{m}], [\frac{2}{m}+\epsilon, \frac{3}{m}], ..., [\frac{m-1}{m}+\epsilon, \frac{m}{m}]$, where $\epsilon$ is a small number (we set $\epsilon = \frac{1}{10m}$). Binary decision variable $v_{i,p}$ equals 1, if the value of feature $i \in \mathcal{I}$ falls into range $p \in \mathcal{P}= \{1, 2, ..., m\}$.
The master problem is formulated on the basis of PMO model \eqref{model:PMO} with additional constraints connecting decision variables $\boldsymbol{x}$ and $v_{i,p}$: 
\begin{subequations}
\begin{align}
& 0 \leq x_i \leq \frac{1}{m} + 1 - v_{i,1}  &\forall i \in \mathcal{I}  \tag{B.1}\label{constraint:SVM-BC1}\\
& (\frac{p-1}{m}+\epsilon)v_{i,p} \leq x_i \leq \frac{p}{m} + 1 - v_{i,p}  &\forall i \in \mathcal{I},\ \forall p \in \mathcal{P}\ \text{and} \ p \geq 2 \tag{B.2}\label{constraint:SVM-BC2}\\
& \sum_{p \in \mathcal{P}} v_{i,p} = 1  &\forall i \in \mathcal{I} \tag{B.3}\label{constraint:SVM-BC3} \\ 
& v_{i,p} \in \{0, 1\}  &\forall i \in \mathcal{I},\ \forall p \in \mathcal{P} \tag{B.4}\label{constraint:SVM-BC4} 
\end{align}
\end{subequations}

We describe the procedure of our proposed branch-and-cut algorithm in Algorithm \ref{algo:SVM-BC} and note that this branch-and-cut algorithm is a heuristic. The algorithm can be implemented using a call-back function by dynamically checking  sub-problem constraints for integer nodes, and adding cuts while the branch-and-bound tree is being built.

\begin{algorithm}
\begin{algorithmic}
\caption{The Branch-and Cut Procedure for PMO with One-Class SVM Constraints}\label{algo:SVM-BC}
\State Candidate solution $(\boldsymbol{\bar{x}},\ \boldsymbol{\bar{v}}) \gets \text{solve PMO \eqref{model:PMO} with constraints \ref{constraint:SVM-BC1}-\ref{constraint:SVM-BC4}}$
\If{$\boldsymbol{\bar{x}}$ violates SVM constraints \ref{constraint:SVM1}} 
    \State Add a cover cut $\sum_{i \in \mathcal{I}}\sum_{p \in \mathcal{P}: \bar{v}_{i,p}=1} v_{i,p} \leq n-1$
\Else
    \State \Return $\boldsymbol{\bar{x}}$ as the final solution
\EndIf 

\end{algorithmic}
\end{algorithm}

Note that the cover cuts ensure that any given combination of the discretization ranges is explored at most once during the execution of the algorithm. The algorithm terminates in a finite number of steps because there is a finite number of combinations for the discretization ranges.

\newpage

\section{Running Time Summary of Each Benchmark Function} \label{sec:appendix-running time}
\begin{figure}[h]
	\centering
	\captionsetup{width=\linewidth}
	\includegraphics[scale=0.58]{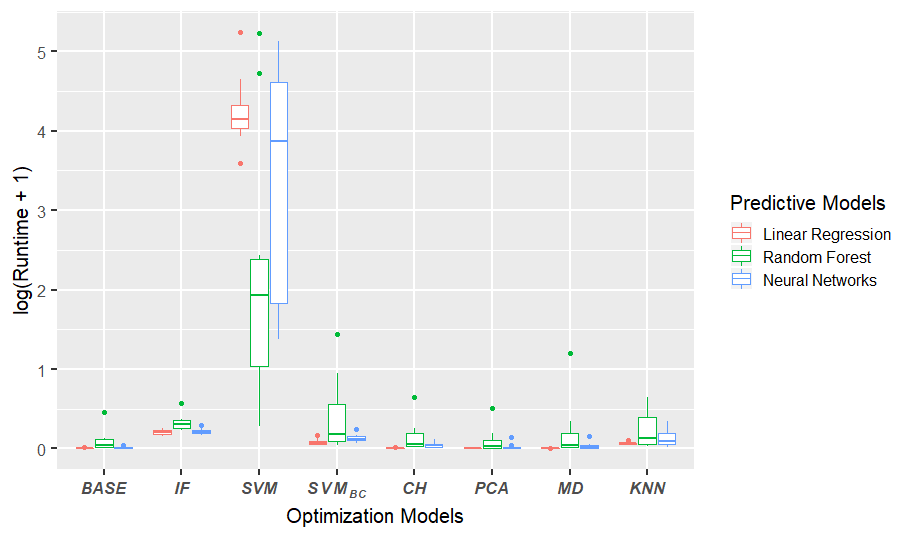}
	\caption{Solution times of instances generated with \textbf{Beale} and three predictive models using different optimization models in natural logarithmic scale}
	\label{fig:runtime_Beale}
\end{figure}

\begin{figure}[h]
	\centering
	\captionsetup{width=\linewidth}
	\includegraphics[scale=0.58]{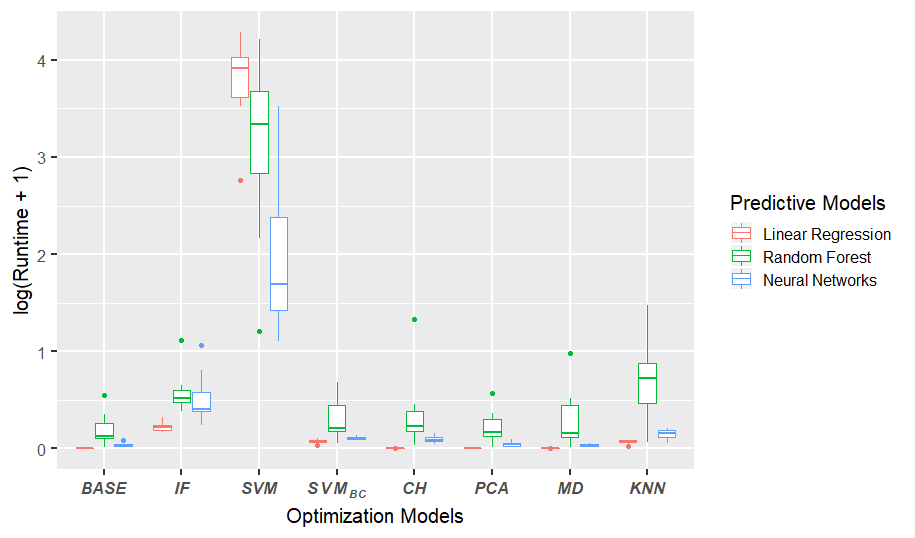}
	\caption{Solution times of instances generated with \textbf{Peaks} and three predictive models using different optimization models in natural logarithmic scale} 
	\label{fig:runtime_Peaks}
\end{figure}

\begin{figure}[h]
	\centering
	\captionsetup{width=\linewidth}
	\includegraphics[scale=0.58]{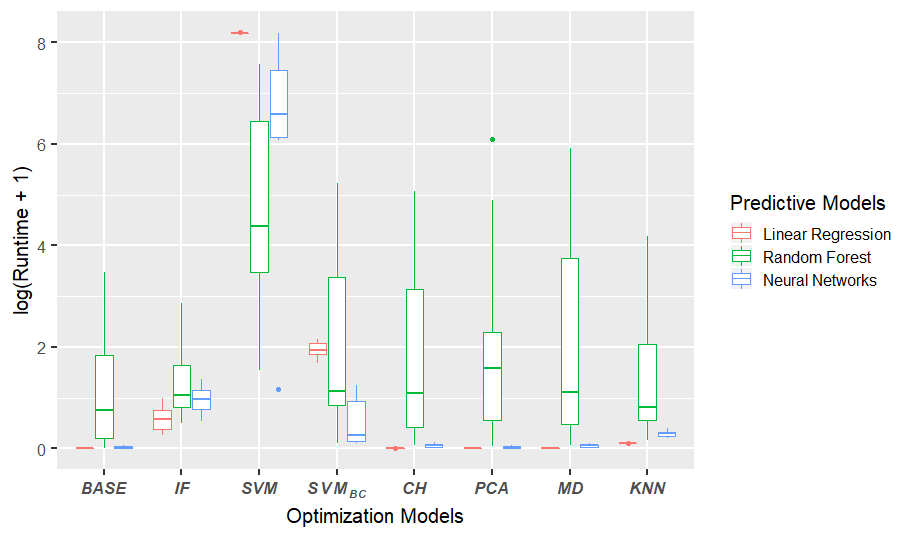}
	\caption{Solution times of instances generated with \textbf{Griewank} and three predictive models using different optimization models in natural logarithmic scale}
	\label{fig:runtime_Griewank}
\end{figure}

\begin{figure}[h]
	\centering
	\captionsetup{width=\linewidth}
	\includegraphics[scale=0.58]{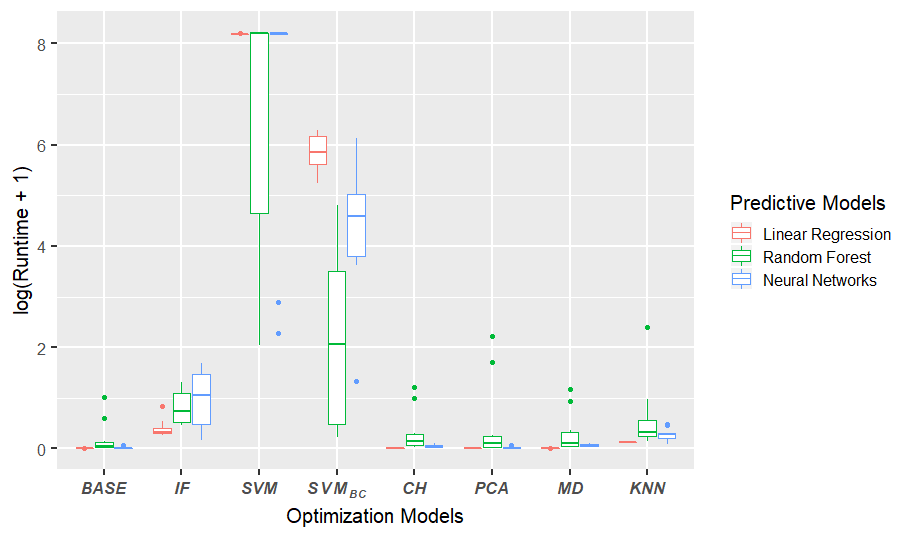}
	\caption{Solution times of instances generated with \textbf{Quintic} and three predictive models using different optimization models in natural logarithmic scale}
	\label{fig:runtime_Quintic}
\end{figure}

\begin{figure}[h]
	\centering
	\captionsetup{width=\linewidth}
	\includegraphics[scale=0.58]{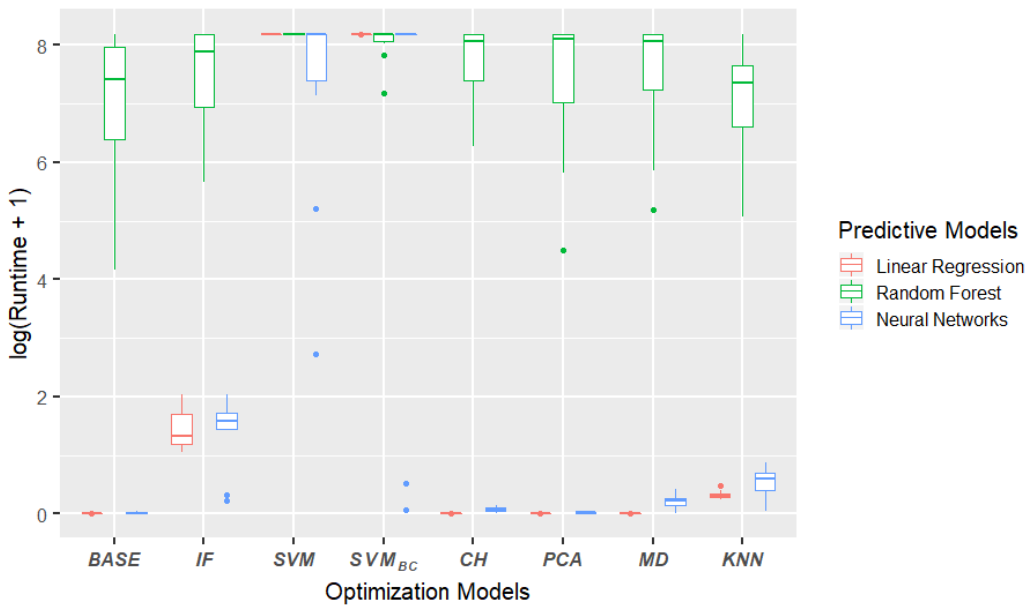}
	\caption{Solution times of instances generated with \textbf{Rastrigin} and three predictive models using different optimization models in natural logarithmic scale}
	\label{fig:runtime_Rastrigin}
\end{figure}
\end{appendices}

\end{document}